\RequirePackage{fix-cm}
\documentclass[twocolumn]{svjour3}          
\smartqed  
\usepackage{graphicx}
\usepackage{lineno,hyperref}
\usepackage{epsfig}
\usepackage{graphicx}
\usepackage{epsfig}
\usepackage{amsmath}
\usepackage{amssymb}
\usepackage{multirow}
\usepackage[normalem]{ulem}
\usepackage{comment}
\usepackage{sidecap}
\usepackage{xcolor}
\usepackage{booktabs}
\usepackage{tabularx}
\usepackage{algorithm}
\usepackage[noend]{algpseudocode}

\usepackage{subfig}

\DeclareMathOperator*{\argmin}{arg\,min}

\DeclareMathOperator{\Acc}{Acc}
\DeclareMathOperator{\E}{E}
\DeclareMathOperator{\Var}{Var}

\newcommand{\be}{\mathbf{e}}

\newcommand{\bx}{\mathbf{x}}

\newcommand{\sE}{\mathcal{E}}

\newcommand{\sX}{\mathcal{X}}
\newcommand{\sY}{\mathcal{Y}}
\newcommand{\sZ}{\mathcal{Z}}

\newcommand{\bbR}{\mathbb{R}}

\newcommand{\figref}[1]{Figure~\ref{#1}}

\newcommand{\tabref}[1]{Table~\ref{#1}}

\usepackage{pifont}
\newcommand{\cmark}{\text{\ding{51}}}
\newcommand{\xmark}{\text{\ding{55}}}

\begin{document}

\title{Zero-Shot Learning on 3D Point Cloud Objects and Beyond}


\author{Ali~Cheraghian \and Shafin~Rahman \and Townim~F.~Chowdhury \and Dylan~Campbell \and Lars~Petersson}


\institute{	  Ali Cheraghian \at
			  Data61, CSIRO, ACT 2601, AU \\
              Australian National University, Canberra ACT 0200 AU \\
              \email{ali.cheraghian@anu.edu.au}
           \and
           	  Shafin Rahman \at
              North South University, Dhaka, Bangladesh\\
              \email{shafin.rahman@northsouth.edu}
           \and
           	  Townim F. Chowdhury \at
              North South University, Dhaka, Bangladesh\\
              \email{townim.faisal@northsouth.edu}          
            \and
              Dylan Campbell \at
              University of Oxford, Oxford, United Kingdom \\
              \email{dylan@robots.ox.ac.uk} 
            \and
              Lars Petersson \at
              Data61, CSIRO, ACT 2601, AU \\
              \email{lars.petersson@data61.csiro.au}               
}

\date{Received: date / Accepted: date}

\maketitle

\begin{abstract}
Zero-shot learning, the task of learning to recognize new classes not seen during training, has received considerable attention in the case of 2D image classification. However, despite the increasing ubiquity of 3D sensors, the corresponding 3D point cloud classification problem has not been meaningfully explored and introduces new challenges. In this paper, we identify some of the challenges and apply 2D Zero-Shot Learning (ZSL) methods in the 3D domain to analyze the performance of existing models. Then, we propose a novel approach to address the issues specific to 3D ZSL. We first present an inductive ZSL process and then extend it to the transductive ZSL and Generalized ZSL (GZSL) settings for 3D point cloud classification. To this end, a novel loss function is developed that simultaneously aligns seen semantics with point cloud features and takes advantage of unlabeled test data to address some known issues (e.g., the problems of domain adaptation, hubness, and data bias). While designed for the particularities of 3D point cloud classification, the method is shown to also be applicable to the more common use-case of 2D image classification. An extensive set of experiments is carried out, establishing state-of-the-art for ZSL and GZSL on synthetic (ModelNet40, ModelNet10, McGill)  and real (ScanObjectNN) 3D point cloud datasets.


%

\end{abstract}

\section{Introduction}


Capturing 3D point cloud data from complex scenes has been facilitated by increasingly accessible and inexpensive 3D depth camera technology. This in turn has expanded the interest in, and need for, 3D object classification methods that can operate on such data. However, much if not most of the data collected will belong to classes for which a classification system may not have been explicitly trained. In order to recognize such previously ``unseen'' classes, it is necessary to develop Zero-Shot Learning (ZSL) methods in the domain of 3D point cloud classification. While such methods are typically trained on a set of so-called ``seen'' classes, they are capable of classifying certain ``unseen'' classes as well. Knowledge about unseen classes is introduced to the network via semantic feature vectors that can be derived from networks pre-trained on image attributes or on a very large corpus of texts~\cite{Hinton_NIPS_2009,Changpinyo_2016_CVPR,Akata_PAMI_2016,Zhang_2017_CVPR,Xian_CVPR_2017}.

Performing ZSL for the purpose of 3D object classification is a more challenging task than ZSL applied to 2D images~\cite{rahman2018unified,Zhang_2017_CVPR,Akata_PAMI_2016,Changpinyo_2016_CVPR,Hinton_NIPS_2009,Lampert_PAMI_2014,Xian_CVPR_2017}. We identify three particular challenges in this regard.

\begin{enumerate}
    \item Availability of high quality pre-trained models: ZSL methods in the 2D domain commonly take advantage of pre-trained models, like ResNet \cite{He2016DeepRL}, that have been trained on millions of labeled images featuring thousands of classes. As a result, the extracted 2D features are very well clustered. By contrast, there is no parallel in the 3D point cloud domain; labeled 3D datasets tend to be small and have only limited sets of classes. For example, pre-trained models like PointNet \cite{Article1} are trained on only a few thousand samples from a small number of classes. This leads to poor-quality 3D features with clusters that are not nearly as well separated as their visual counterparts.
    
    \item The hubness problem: In high-dimensional data, some points---called hubs---occur frequently in the $k$-nearest neighbor sets of other points. This is a consequence of the curse of dimensionality associated with nearest neighbor (NN) search~\cite{Article57}. In ZSL, the hubness problems occurs for two reasons \cite{Shigeto_Hubness_2015}. Firstly, both input and semantic features reside in a high dimensional space. Secondly, ridge regression, which is widely used in ZSL, is known to induce hubness. As a result, it causes a bias in the predictions, with only a few classes predicted most of the time regardless of the query. The hubness problem is exacerbated by the relatively poor quality of 3D features, making it more difficult to relate those features to their corresponding semantics~\cite{Zhang_2017_CVPR}.
    
    \item The domain shift problem: The function learned from seen samples is biased to those samples and cannot generalize well to unseen classes. In the \emph{inductive learning} approach, where only seen classes are used during training, projected semantic vectors tend to move towards the seen feature vectors, making the intra-class distance between corresponding unseen semantic and feature vectors large. Similar to hubness, the domain shift problem is intensified when 
    training is done on seen synthetic 3D point cloud objects (ModelNet40~~\cite{Article10}), but testing on unseen real-world 3D scanned data (ScanObjectNN \cite{scanobjectnn_iccv19}).
\end{enumerate}

Some intuition about these challenges can be attained by visualizing the respective pre-trained feature spaces, as shown in Figure~\ref{fig:2D_versus_3D} for the 3D datasets (a) ModelNet10~\cite{Article10} and (b) ScanObjectNN \cite{scanobjectnn_iccv19}, and the 2D datasets (c) AwA2~\cite{AwA_2009} and (d) CUB~\cite{CUB_2011}.  The quality of the image features is much higher than the point cloud features, with a much more separable cluster structure. When the clusters are not well-separated, the hubness and domain shift problem are worsened. In this paper, we address the following questions for ZSL on 3D point cloud data:

\begin{figure}[t]
\centering
\begin{minipage}[b]{0.45\columnwidth}
  \centering
  \centerline{\includegraphics[width=1\linewidth,trim=0cm 0cm 0cm 0cm, clip]{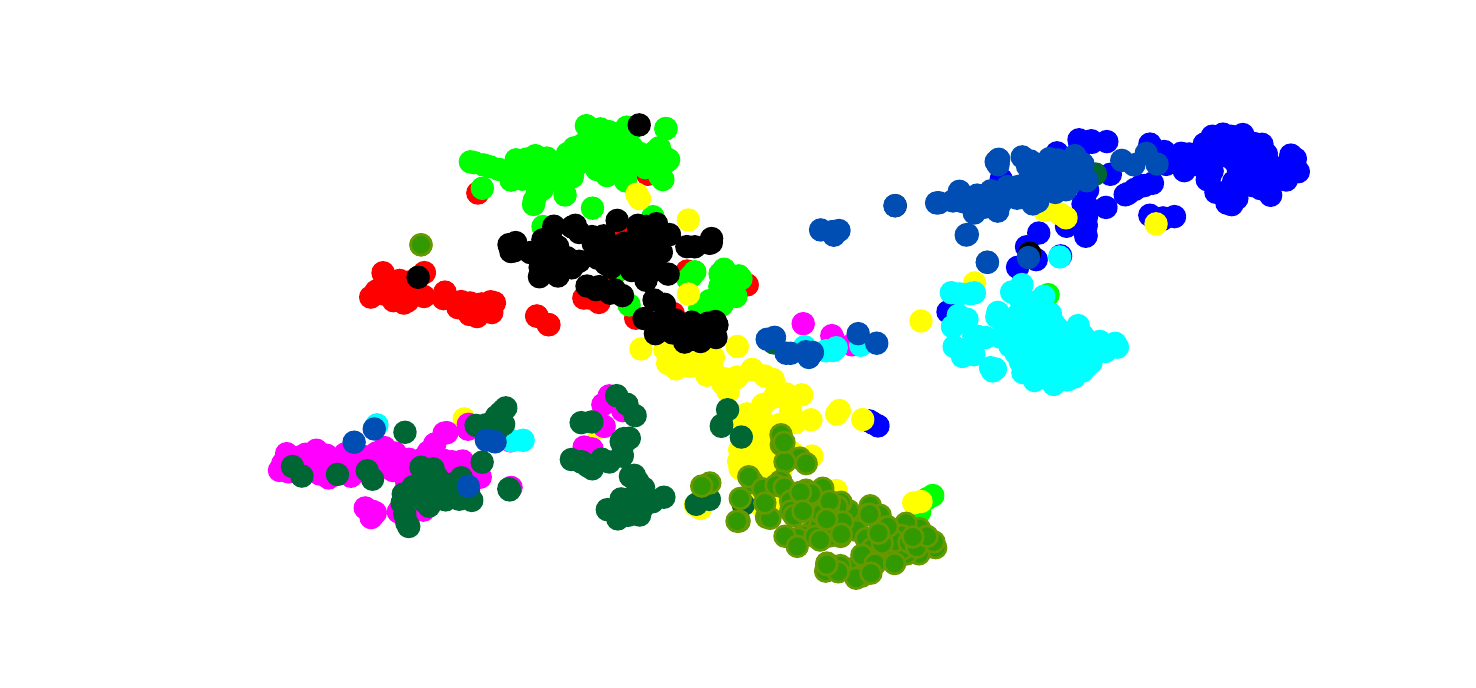}}
  \centerline{\small (a) ModelNet10}\medskip
\end{minipage}
\begin{minipage}[b]{0.5\columnwidth}
  \centering
  \centerline{\includegraphics[width=.5\linewidth,trim=0cm 0cm 0cm 0cm, clip]{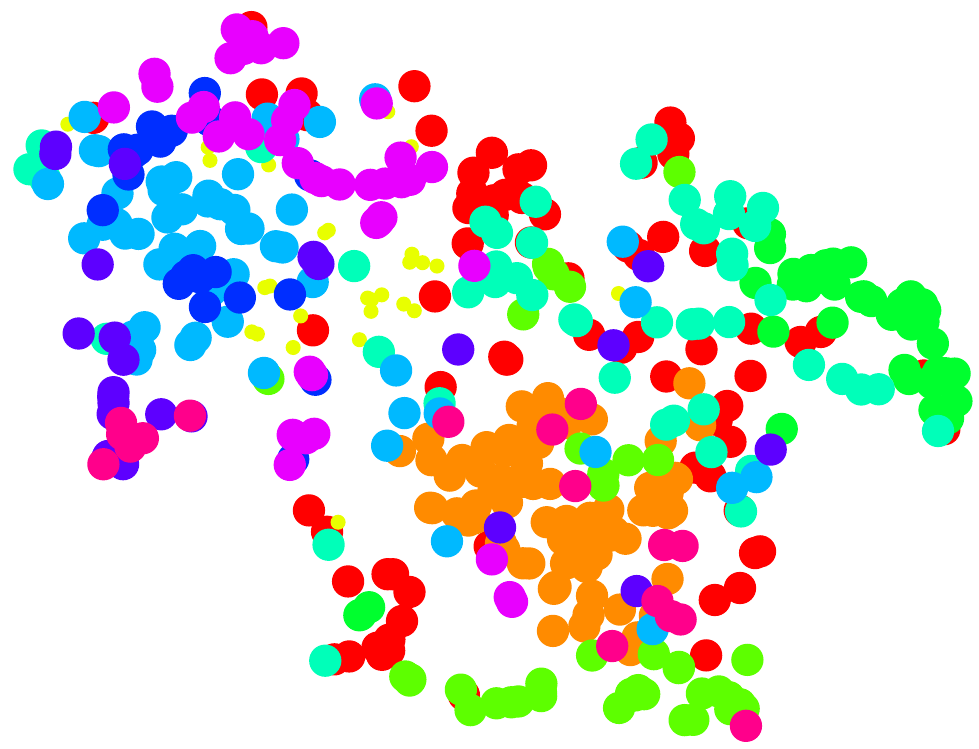}}
  \centerline{\small (b) ScanObjectNN}\medskip
  \vspace{-0mm}
\end{minipage}
\begin{minipage}[b]{0.5\columnwidth}
  \centering
  \centerline{\includegraphics[width=0.5\linewidth,trim=0cm 0cm 0cm 0cm, clip]{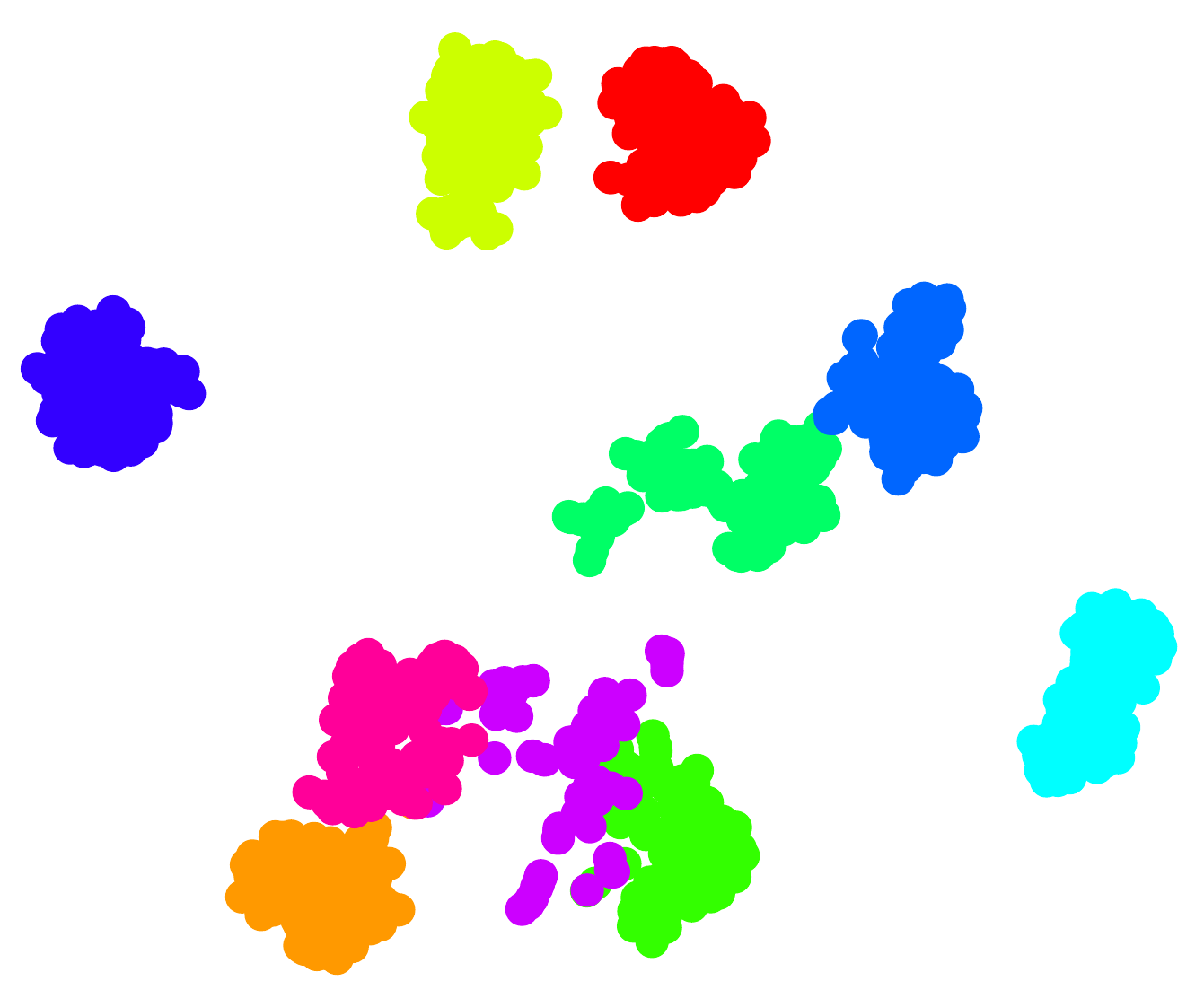}}
  \centerline{\small (c) AwA2 }\medskip
\end{minipage}
\begin{minipage}[b]{0.48\columnwidth}
  \centering
  \centerline{\includegraphics[width=.55\linewidth,trim=0cm 0cm 0cm 1.4cm, clip]{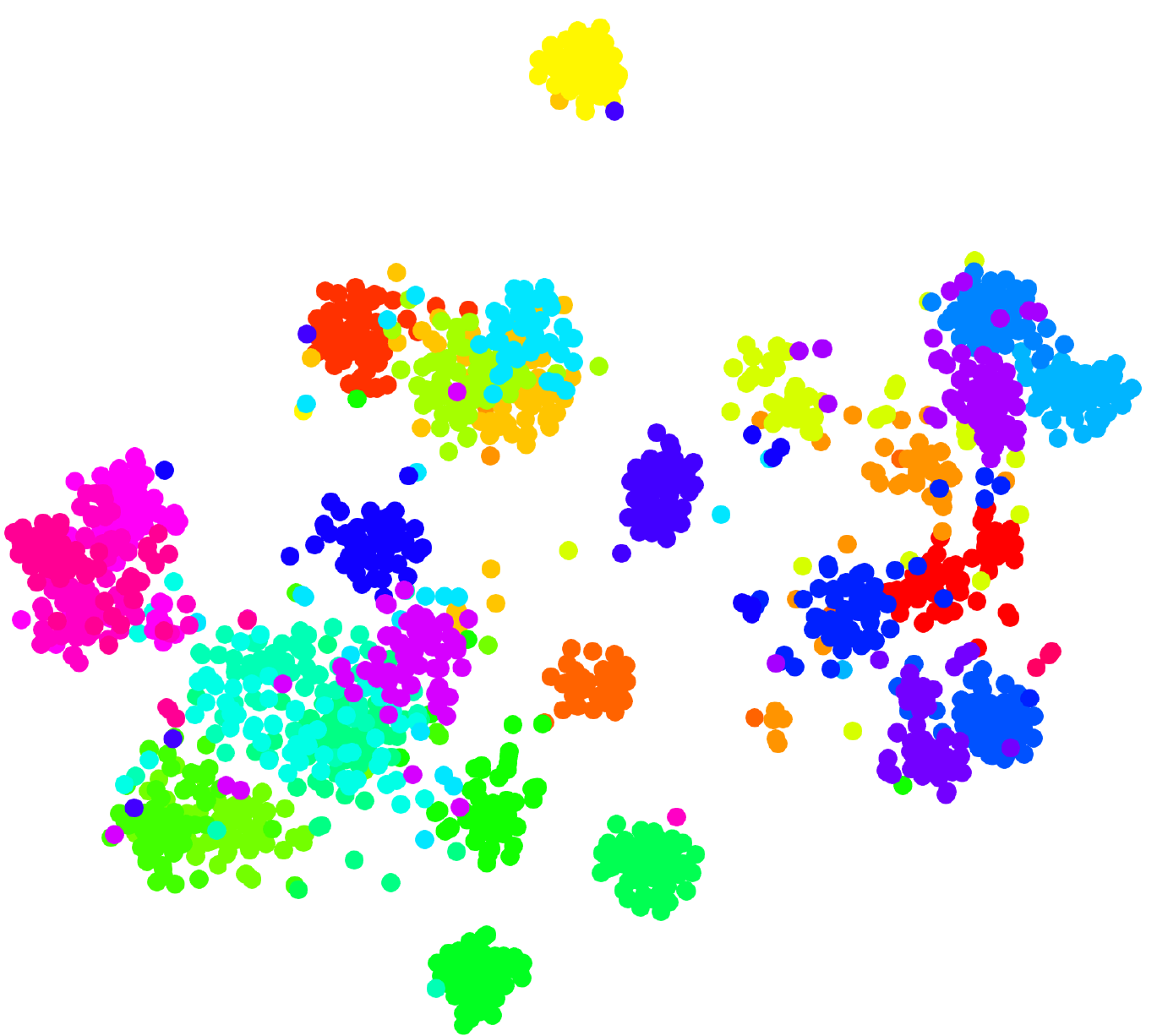}}
  \centerline{\small (d) CUB}\medskip
  
\end{minipage}
\caption{tSNE~\cite{tSNE_van2014} visualizations of unseen 3D point cloud features of the (a) ModelNet10~\cite{Article10} and (b) ScanObjectNN \cite{scanobjectnn_iccv19} datasets, and unseen 2D image features of the (c) AwA2~\cite{Xian_CVPR_2017} and (d) CUB~\cite{CUB_2011} datasets. The cluster structure in the 2D feature space is much better defined, with tighter and more separated clusters than those in the 3D point cloud.}
\label{fig:2D_versus_3D}
\end{figure}

\textit{(a) How do standard ZSL approaches perform on low quality 3D point cloud features?} We conduct a series of experiments utilising four popular structures traditionally used for feature extraction in 3D point clouds. These are PointNet~\cite{Article1},  DGCNN~\cite{Article24}, PointConv~\cite{8954200}, and PointAugment~\cite{Li_2020_CVPR}. 
With the help of these base architectures, we build a structure for ZSL that combines point cloud features with word vector semantic features thereby enabling the classification of previously unseen 3D classes. This combination process follows the standard approach of ZSL that maps the point cloud features to the space of semantic vectors. The performance obtained by this approach shows the complex nature of ZSL tasks on 3D data due to the poor feature quality, the hubness and domain-shift problems. However, it establishes suitable baselines for any ZSL models on 3D point cloud data. 

\textit{(b) How much can the domain shift be mitigated?} In this paper, we attempt to address the domain shift problem using transductive learning. Our goal is to design a strategy that reduces the bias and encourages the projected semantic vectors to align with their true feature vector counterparts, minimizing the average intra-class distance. In 2D ZSL, the transductive setting has been shown to be effective~\cite{Fu_PAMI_2015,Zhao_NIPS_2018,Song2018TransductiveUE}, however, in the case of 3D point cloud data it is a more challenging task. Pre-trained 3D features are poorly clustered and exhibit large intra-class distances. 
In order to take advantage of the transductive learning approach for 3D point cloud zero-shot learning, we propose a transductive ZSL method using a novel triplet loss that is employed in an unsupervised manner. Unlike the traditional triplet formulation \cite{Facenet,BMVC17Zeroshot}, our proposed triplet loss works on unlabeled (test) data and can operate without the need of ground-truth supervision. This loss applies to unlabeled data such that intra-class distances are minimized while also maximizing inter-class distances, reducing the bias problem. In addition to the triplet loss, we also employ a distance-based unbiased loss to balance seen and unseen prediction scores. As a result, a prediction function with greater generalization ability and effectiveness on unseen classes is learned.

\textit{(c) How can we address the hubness problem for 3D data?} The hubness problem occurs when a model is biased to predict a small subset of labels for most of the test instances. Popular ZSL methods on 2D image data usually project the semantic features to the space of visual features to handle the hubness problem. In this paper, we first design our architecture by following the same trend and observe the performance gain of applying the reverse projecting trick. Secondly, to further improve the performance, we propose a new loss for the transductive setting to explicitly alleviate the hubness problem. We calculate this loss by evaluating each unlabeled test data element in an unsupervised manner, and counting the number of times each class gets predicted on the batch. This is used to estimate a measure of hubness: the skewness of the current prediction. We minimize the skewness of each batch to reduce the degree of hubness.

In addition to 3D point cloud data, our proposed method is also applicable in the case of 2D ZSL, which demonstrates the generalization strength of our method to other sensor modalities. Our main contributions are: (1) an evaluation of the zero-shot learning (ZSL) and generalized zero-shot learning (GZSL) tasks for 3D point cloud classification by adapting both inductive and transductive learning settings; (2) a novel triplet loss that takes advantage of unlabeled test data, applicable to both 3D point cloud data and 2D images; (3) an approach to address the hubness and bias problems of G/ZSL in transductive settings; (4) a new evaluation protocol for ZSL methods on 3D point clouds which consists of a seen and unseen split of data from the datasets ModelNet40~\cite{Article10}, ModelNet10~\cite{Article10}, McGill~\cite{Article49} and ScanObjectNN \cite{scanobjectnn_iccv19}, and performing extensive experiments, establishing state-of-the-art on four 3D datasets.

Preliminary sections of this paper have been published previously \cite{Cheraghian_WACV_2020,cheraghian2019mitigating,cheraghian2019zeroshot}. Here, we encapsulate the contributions in a unified framework and extend the previous work as follows: (1) we address the hubness problem in the transductive settings and propose a new loss to balance seen and unseen scores; (2) we analyze the framework in detail, with new ablation studies, and situate it within the context of the related work; (3) we provide extensive evaluation of eight established ZSL and GZSL methods on 3D point cloud data;
and (4) we propose a new seen/unseen split for a real-world scanned 3D object dataset (ScanObjectNN) and evaluate on this dataset.

\section{Related works}

\noindent\textbf{3D point cloud object recognition architecture:} The early methods utilizing deep learning for operating on 3D point clouds used volumetric~\cite{Article10} or multi-view~\cite{Article13} representations in order to work with 3D data. Recently, the trend in this area has shifted to instead using raw point clouds directly~\cite{Article2,Article24,Article27}, without any preprocessing step. These methods do not suffer to the same degree from scalability issues as the volumetric representation does, and they do not make any {\it a priori} assumptions onto which 2D planes, and how many, that the point cloud should be projected on, like the view-based methods do. PointNet~\cite{Article1} was the first work that operated on raw point clouds directly at the input of the network. PointNet used a multi-layer perceptron (mlp)~\cite{Article42} to extract features from point sets, and max-pooling layers to remove the otherwise inherent issue of permutation from the point clouds. Later, many methods~\cite{Article2,Article24,Article27,8954200,Li_2020_CVPR,8658405} were proposed to overcome the limitations of PointNet, which does not utilize local features or a more advanced pooling operation than max-pooling. The traditional recognition where all the classes of interest have been seen at training time, have been considered in the case of 3D point cloud data. The current literature does not fully address the zero-shot version of the 3D recognition problem \cite{Cheraghian_WACV_2020,cheraghian2019mitigating,cheraghian2019zeroshot}. In this paper, we perform both transductive and inductive ZSL and GZSL on 3D point cloud objects. 

\noindent\textbf{Zero-Shot Learning:} For the ZSL task, there has been significant progress, including on image recognition~\cite{rahman2018unified,Zhang_2017_CVPR,Akata_PAMI_2016,Changpinyo_2016_CVPR,Hinton_NIPS_2009,Lampert_PAMI_2014,Xian_CVPR_2017}, multi-label ZSL~\cite{Lee_2018_CVPR,rahman2018deep}, and zero-shot detection~\cite{rahman2018ZSD}. Despite this progress, these methods solve the constrained problem where the test instances are restricted to only unseen classes, rather than being from either seen or unseen classes. This setting, where both seen and unseen classes are considered at test time, is called Generalized Zero-Shot Learning (GZSL). To address this problem, some methods decrease the scores that seen classes produce by a constant value \cite{Chao_ECCV_2016}, while others perform a separate training stage intended to balance the probabilities of the seen and unseen classes \cite{rahman2018unified}. Also, some Generative Adversarial Networks (GAN) based approaches~\cite{Xian_2018_CVPR,gdan-cvpr19,Schonfeld_2019_CVPR,lisgan-cvpr19,tfvaegan-eccv20,lsrgan-eccv20,bidirectional-gan-2020,zerovaegan-20} have been proposed to solve ZSL and GZSL problems in recent years. Schonfeld \textit{et al.} \cite{Schonfeld_2019_CVPR} learned a shared latent space of image features and semantic representation based on a modality-specific VAE model. In our work, we propose novel loss functions (for both inductive and transductive cases) to address the bias problem, leading to significantly better GZSL results.

\noindent\textbf{Transductive Zero-shot Learning:} The transductive learning approach takes advantage of unlabeled test samples, in addition to the labeled seen samples. For example, 
Rohrbach \textit{et al.} \cite{Rohrbach_NIPS_2013} exploited the manifold structure of unseen classes using a graph-based learning algorithm to leverage the neighborhood structure within unseen classes. Fu \textit{et al.} \cite{Fu_PAMI_2015} proposed a multi-view transductive setting to address projection shift and to exploit various semantic representations of the visual feature. 
Yu \textit{et al.} \cite{Yu_TCy_2018} proposed a transductive approach to predict class labels via an iterative refining process. Guo \textit{et al.} \cite{Guo_AAAI_2016} proposed a joint learning method that learns a shared model space to share knowledge between seen and unseen classes using semantic attributes jointly. All of these methods attempt to improve the accuracy of the unseen classes in transductive settings. More recently, transductive ZSL methods have started exploring how to improve the accuracy of both the seen and unseen classes in generalized ZSL tasks \cite{Zhao_NIPS_2018,Song2018TransductiveUE}. Zhao \textit{et al.} \cite{Zhao_NIPS_2018} proposed a domain invariant projection method that projects visual features to semantic space and reconstructs the same feature from the semantic representation in order to narrow the domain gap. In another approach, Song \textit{et al.} \cite{Song2018TransductiveUE} identified the model bias problem of inductive learning, that is, a trained model assigns higher prediction scores for seen classes than unseen. To address this, they proposed a quasi-fully supervised learning method to solve the GZSL task. Xian \textit{et al.} \cite{Xian_2019_CVPR} proposed f-VAEGAN-D2 which takes advantage of both VAEs and GANs to learn the feature distribution of unlabeled data. Narayan \textit{et al.} \cite{tfvaegan-eccv20} followed the same setting as proposed in the baseline f-VAEGAN-D2 \cite{Xian_2019_CVPR}. Gao \textit{et al.} \cite{zerovaegan-20} used K-Nearest Neighbors and classification probability to provide pseudo-labels for unlabeled unseen features. All of these approaches are designed for transductive ZSL tasks on 2D image data. In contrast, we explore to what extent a transductive ZSL setting helps to improve 3D point cloud recognition.

\noindent\textbf{The Hubness Problem:} The hubness problem in high dimensional nearest neighbor search spaces was first investigated in \cite{Article57} where they illustrate that the hubness problem is related to the data distribution in the high dimensional space. In later studies \cite{article56,Shigeto_Hubness_2015,Zhang_2017_CVPR}, the hubness problem in ZSL is investigated. Dinu \textit{et al.}~\cite{article56} proposed an algorithm that corrects the hubness problem by using more unlabeled seen data in addition to test instances. Shigeto \textit{et al.}~\cite{Shigeto_Hubness_2015} mentioned that the projection function used for least squares regularization affect the hubness problem negatively and instead introduces a reverse regularized function in order to weaken the hubness problem. In contrast to the mentioned works, Zhang \textit{et al.}~\cite{Zhang_2017_CVPR} proposed to deal with the hubness problem by instead considering the feature space as the embedding space. In this paper, we address the hubness problem of ZSL on 3D point cloud classification.  

\noindent{\textbf{Learning with a Triplet Loss:}} Triplet losses have been widely used in computer vision \cite{Facenet,BMVC17Zeroshot,Dong_2018_ECCV,He_2018_CVPR,Do_2019_CVPR}. Schroff \textit{et al.} \cite{Facenet} demonstrated how to select positive and negative anchor points from visual features within a batch. Qiao \textit{et al.} \cite{BMVC17Zeroshot} introduced using a triplet loss to train an inductive ZSL model. More recently, Do \textit{et al.} \cite{Do_2019_CVPR} proposed a tight upper bound of the triplet loss by linearizing it using class centroids, Zakharov \textit{et al.} \cite{8202207} explored the triplet loss in manifold learning, Srivastava \textit{et al.} \cite{article1323} investigated weighting hard negative samples more than easy negatives, and Zhaoqun \textit{et al.} \cite{Li2018AngularTL} proposed the angular triplet-center loss, a variant that reduces the similarity distance between features. Triplet loss related methods typically work under inductive settings, where the ground-truth label of an anchor point remains available during training. In contrast, we describe a triplet formation technique in the transductive setting. Our method utilizes test data without knowing its true label. Moreover, we choose positive and negative samples of an anchor from word vectors instead of features.

\section{Zero-Shot Learning for 3D Point Clouds}




The comparative lack of large-scale 3D datasets with many object categories has meant that 3D features are not as robust and separable as 2D features. As a result, relating 3D features to their corresponding semantic vectors is more difficult than for the 2D case. Addressing the poor feature quality of typical 3D datasets, we investigate suitable 3D point cloud architectures and loss functions in both transductive and inductive settings. Our method specifically addresses the alignment of poor features (like those coming from 3D feature extractors) with semantic vectors. Therefore, while our method improves the results for both 2D and 3D modalities, the largest gain is observed in the 3D case.


\subsection{Problem formulation}
Let $\sX = \{\bx_{i}\}_{i = 1}^{n}$ for $\bx_{i}\in\bbR^{3}$ denote a 3D point cloud. Also let $\sY^{s} = \{y_{i}^{s}\}_{i = 1}^{S}$ and $\sY^{u} = \{y_{i}^{u}\}_{i = 1}^{U}$ denote disjoint ($\sY^{s}\cap\sY^{u}=0$) seen and unseen class label sets with sizes $S$ and $U$ respectively, and $\sE^{s}=\{\phi(y^{s}_{i})\}_{i = 1}^{S}$ and $\sE^{u}=\{\phi(y^{u}_{i})\}_{i = 1}^{U}$ denote the sets of associated semantic embedding vectors for the embedding function $\phi(\cdot)$, with $\phi(y)\in\bbR^{d}$.
Then we define the set of $n_{s}$ seen instances as $\sZ^{s} = \{(\sX_{i}^{s}, l_{i}^{s}, \be_{i}^{s})\}_{i=1}^{n_{s}}$, where $\sX_{i}^{s}$ is the $i$\textsuperscript{th} point cloud of the seen set with label $l_{i}^{s} \in \sY^{s}$ and semantic vector $\be_{i}^{s} = \phi(l_{i}^{s}) \in \sE^{s}$.
The set of $n_{u}$ unseen instances is defined similarly as $\sZ^{u} = \{(\sX_{i}^{u},l_{i}^{u},\be_{i}^{u})\}_{i=1}^{n_{u}}$, where $\sX_{i}^{u}$ is the $i$\textsuperscript{th} point cloud of the unseen set with label $l_{i}^{u} \in \sY^{u}$ and semantic vector $\be_{i}^{u} = \phi(l_{i}^{u}) \in \sE^{u}$.

We consider two learning problems in this work: zero-shot learning and its generalized variant. The goal of each problem is defined as follows.
\begin{itemize}
  \item Zero-Shot Learning (ZSL): predict a class label $\hat{y}^{u} \in \sY^{u}$ from the unseen label set given an unseen point cloud $\sX^{u}$.
  \item Generalized Zero-Shot Learning (GZSL): predict a class label $\hat{y} \in \sY^{s}\cup\sY^{u}$ from the seen or unseen label sets given a point cloud $\sX$.
\end{itemize}

In this paper, we solve ZSL and GZSL problems in both the inductive and transductive setting. Transductive settings allow the use of unlabeled unseen point cloud instances $\sX^{u}$ during the training stage, whereas inductive settings do not allow access to this unlabeled information.

\subsection{Point cloud feature extractors}

\begin{figure}
\centering
\includegraphics[width=1\linewidth,trim=0cm 0cm 0cm 0cm, clip]{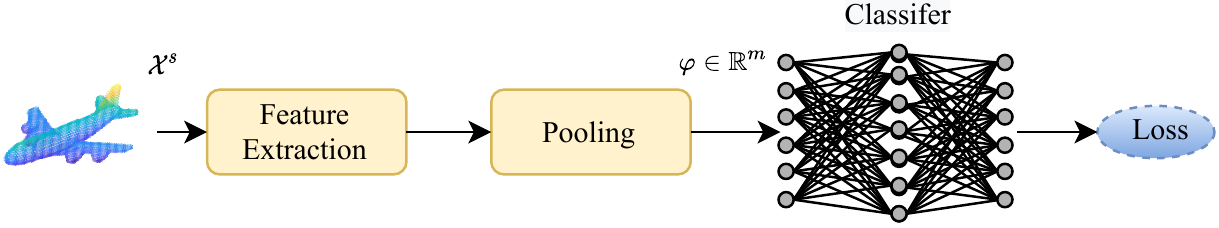}
\caption{General framework of a point cloud architecture. A traditional 3D point cloud recognition system consists of a feature extraction module, a pooling module, and a classifier. We design our backbone using such frameworks.}
\label{fig:arch_point_cloud}
\end{figure}


Given an unordered point set representing an object from a seen class $\sX^{s} =\left \{ \mathbf{x}_{1}^{s},...,\mathbf{x}_{n}^{s} \right \}$, a set function is defined such that any permutation of the point set is irrelevant,
\begin{align*}
f(\mathbf{x}_{1}^{s},\mathbf{x}_{2}^{s},...,\mathbf{x}_{n}^{s}) \approx  g(h(\mathbf{x}_{1}^{s},\beta  )),h(\mathbf{x}_{2}^{s},\beta  )),...,h(\mathbf{x}_{n}^{s},\beta  ))
\end{align*}
where $f$ is the set function, $h$ is the feature extraction function, $g$ is the pooling function with the ability to remove the effects of permutation of points in a set, and $\beta$ represents a set of arguments associated with $\mathbf{x}_i^{s}$. The feature extraction function $h(\mathbf{x}_i^{s},\beta )$ extracts a richer representation from the point cloud in a higher dimension. 
For instance, in PointNet~\cite{Article1}, $h(\mathbf{x}_{i}^{s},\beta )=h(\mathbf{x}_{i}^{s}) : \mathbb{R}^{d}\rightarrow\mathbb{R}^{d^{\prime}},\beta=\left \{ \emptyset \right \} $, since each point is considered separately, the extracted feature vector contains global information. As another example, in DGCNN~\cite{Article24}, which extracts local features as well as global features, $h(\mathbf{x}_{i}^{s},\beta)=h(\mathbf{x}_{i}^{s},\mathbf{x}_{j}^{s}-\mathbf{x}_{i}^{s}) : \mathbb{R}^{d}\times \mathbb{R}^{d}\rightarrow\mathbb{R}^{d^{\prime}},\beta =\left \{ \mathbf{x}_{j}^{s}-\mathbf{x}_{i}^{s} \right \} $. In this case, point sets are represented by a dynamic graph and edge features based on $k$-nearest neighbors are calculated. Since point sets are inherently unordered, a function which is invariant to permutation is necessary to pool point features into a feature vector. Here, $g$, is capable of removing the effects of permutation from point clouds. 
Finally, via a collection of $h(\mathbf{x}_i^{s},\beta )$, corresponding values of $f$ can be computed to form a vector $\varphi(\sX^{s}) \in \bbR^{m}$ 
The obtained feature vector removes permutation from the point cloud. In the next step, a few fully-connected layers are applied to the feature vector $\varphi(\sX^{s})$ in order to transform the features into label space, where a cross-entropy loss is used to train the point cloud backbone. 
We illustrate the point cloud feature extractor architecture in Figure \ref{fig:arch_point_cloud}. 




\begin{figure*}
\centering
\includegraphics[width=1\linewidth,trim=0cm 0cm 0cm .2cm, clip]{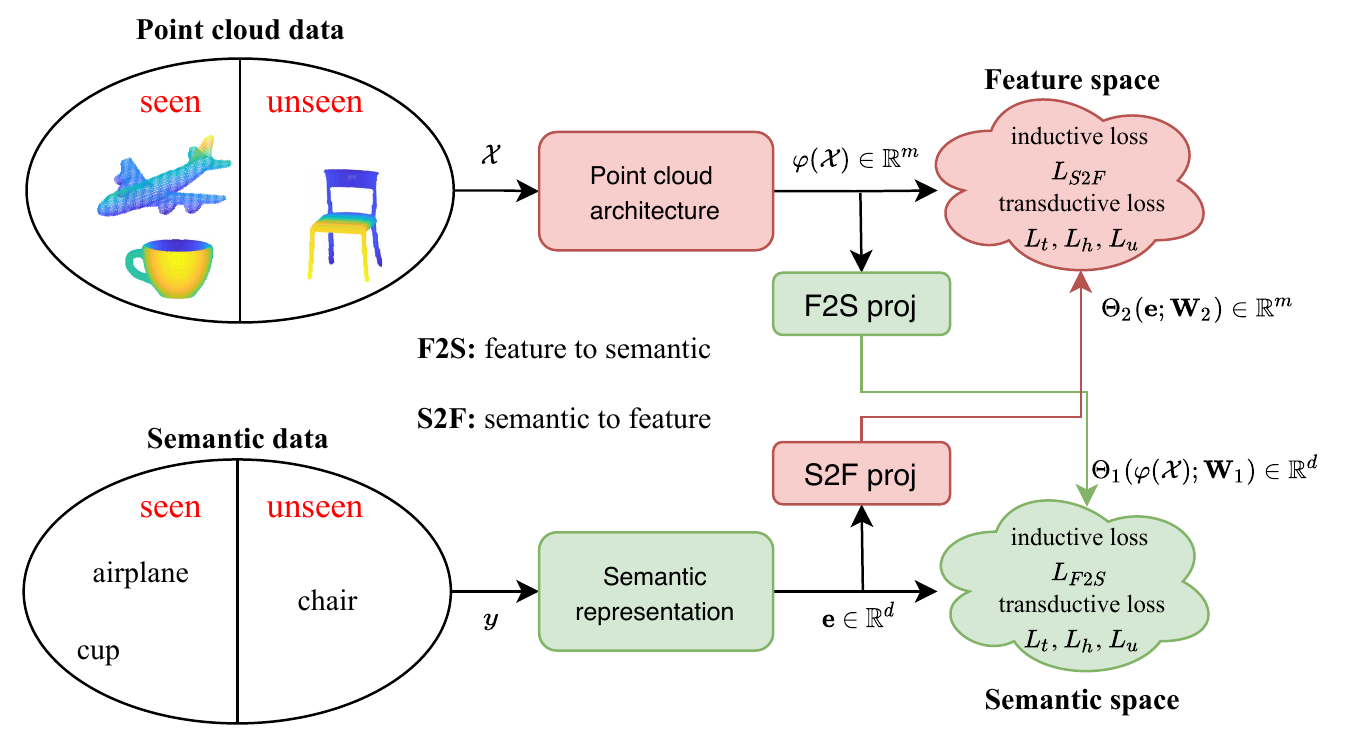}
\caption{The proposed architecture for ZSL and GZSL. For inductive learning, the input point cloud and semantic representation are $\sX = \sX^{s}$ and $\be = \phi(y) \in \sE^{s}$, respectively. For transductive learning, the input point cloud and semantic representation are $\sX = \sX^{s} \cup \sX^{u}$ and $\be \in \sE^{s} \cup \sE^{u}$ respectively. We project point cloud features to semantic space (F2S) or semantic vectors to feature space (S2F) and calculate distances between feature and semantics. Our proposed losses minimize those distances in both the inductive and transductive setting.
}
\label{fig:model_architecture}
\end{figure*}

\subsection{Inductive ZSL on point cloud data}

Our model is trained in a fully-supervised manner with seen instances only from the set $\sZ^{s}$. Let $N$ be the number of instances in the batch and $\varphi(\sX_{i}^{s}) \in \bbR^{m}$ be the point cloud feature vector associated with point cloud $\sX_{i}^{s}$. For ZSL, both point cloud feature $\varphi(\sX)$ and semantic $\sE$ vectors need to embed into the same embedding space. In the ZSL literature, this is done in two ways, and we investigate both in the context of 3D point cloud objects. 


\noindent\textbf{Feature to Semantic (F2S):} a point cloud feature $\varphi(\sX_{i}^{s})$ is projected into the semantic embedding space $\sE$ using a nonlinear projection function $\Theta_1(\cdot)$ with weights $W_1$. The network calculates the following loss:
\begin{align}
L_{F2S}=\frac{1}{N}\sum_{i=1}^{N} \left\| \Theta_1 (\varphi (\sX_{i}^{s}); W_1)-\be_{i}^{s} \right \|_{2}^{2} + \lambda_1 \left \| W_1 \right \|_{2}^{2}
\label{eqn:F2S}
\end{align}
where the parameter $\lambda_1$ controls the amount of regularization.

\noindent\textbf{Semantic to Feature (S2F):} a semantic vector $\be_i^s$ is projected into point cloud feature space using the nonlinear projection function $\Theta_2(\cdot)$ weights $W_2$. The network calculates the following loss:
\begin{align}
L_{S2F}=\frac{1}{N}\sum_{i=1}^{N} \left\| \varphi (\sX_{i}^{s})-\Theta_2 (\be_{i}^{s}; W_2) \right \|_{2}^{2} + \lambda_2 \left \| W_2 \right \|_{2}^{2}
\label{eqn:S2F}
\end{align}
where the parameter $\lambda_2$ controls the amount of regularization.

Zhang \textit{et al.} \cite{Zhang_2017_CVPR} argue that ZSL models based on Semantic to Feature (S2F) projection exhibit less hubness than Feature to Semantic (F2S) projection models. In our experiments, we add evidence that this is also true for 3D data. For the remainder of this treatment, we follow S2F embedding (that is, projection with $\Theta_2(.)$) for ZSL. 




\subsection{Transductive ZSL on point cloud data}

\begin{figure}[!t]\centering
\includegraphics[width=1\linewidth,trim=2cm .1cm 1cm .8cm, clip]{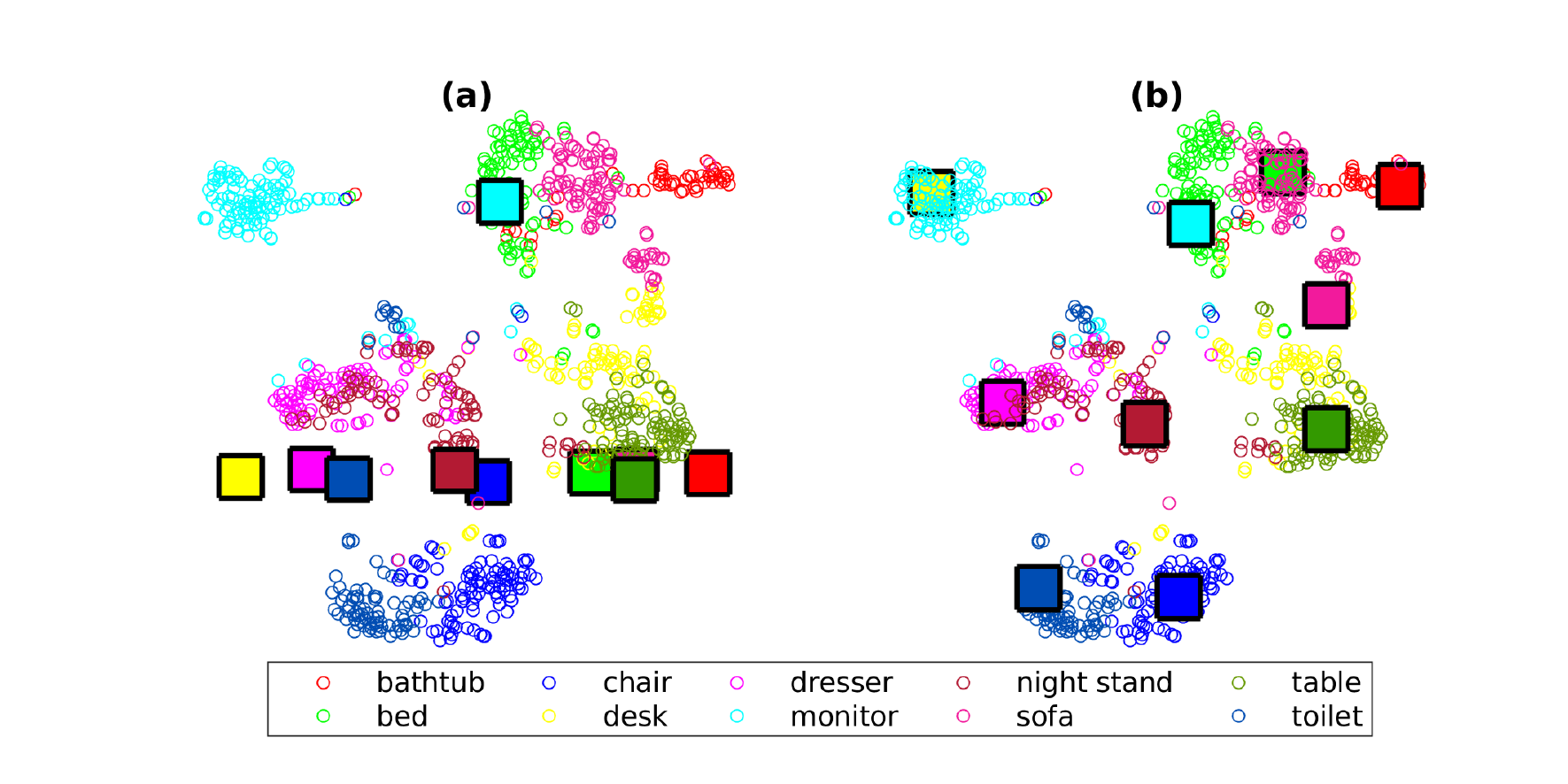}
\caption{2D tSNE~\cite{tSNE_van2014} visualization of unseen point cloud feature vectors (circles) and projected semantic feature vectors (squares) based on (a) inductive and (b) transductive learning on ModelNet10. The projected semantic feature vectors are much closer to the cluster centres of the point cloud feature vectors for transductive ZSL than for inductive ZSL, showing that the transductive approach is able to narrow the domain gap between seen and unseen classes.
}
\label{fig:inductive_transductive}
\vspace{-4mm}
\end{figure}

Transductive ZSL addresses the problem of the projection domain shift~\cite{Fu_PAMI_2015} inherent in inductive ZSL approaches. In ZSL, the seen and unseen classes are disjoint and often only very weakly related. Since the underlying distributions of the seen and unseen classes may be quite different, the ideal projection function between the semantic embedding space and point cloud feature space is also likely to be different for seen and unseen classes. 
As a result, using the projection function learned from only the seen classes without considering the unseen classes will cause an unknown bias.
Transductive ZSL reduces the domain gap and the resulting bias by using unlabeled unseen class instances during training, improving the generalization performance. The effect of the domain shift in ZSL is shown in Figure~\ref{fig:inductive_transductive}.
When inductive learning is used (a), the projected unseen semantic embedding vectors are far from the cluster centres of the associated point cloud feature vectors, however, when transductive learning is used (b), the vectors are much closer to the cluster centres.

\noindent\textbf{Unsupervised triplet loss:} We propose an unsupervised triplet loss that operates on the unlabeled test data.
To compute a triplet loss, a positive and negative sample need to be found for each anchor sample \cite{Facenet}. In the fully-supervised setting, selecting positive and negative samples is not difficult, because all training samples have ground-truth labels. However, it is more challenging in the unsupervised setting, where ground-truth labels are not available. For transductive ZSL, we define a positive sample using a \emph{pseudo-labeling} approach \cite{Pseudo_label}. For each anchor $\sX^{u}$, we assign a pseudo-label that chooses a positive sample $\be^{+}$
among the semantic embedding vectors which is the closest to the anchor feature vector $\varphi(\sX^{u})$ after projection $\Theta_2(\cdot)$, as follows
\begin{align}
\be^{+} = \argmin_{\be \in \sE^{u}} \| \varphi (\sX^{u})-\Theta_2 (\be; W_2) \|_{2}^{2} \textrm{.}\label{arg_ZSL}
\end{align}

Such pseudo-labeling is different from the usual practice \cite{Pseudo_label} because it chooses a semantic vector as a positive sample in the triplet formation instead of a plausible ground-truth label. For GZSL, the unlabeled data $\sX^{c}$ for $c \in \{s, u\}$ can be from the seen or unseen classes during training. As a result, a pseudo-label must be found for both unlabeled seen and unlabeled unseen samples. Importantly, if the pseudo-label indicates that an unlabeled sample is from a seen class, then that sample is discarded. This reduces the impact of incorrect, noisy pseudo-labels on the model for seen classes. Samples from seen classes (with ground-truth labels) will instead influence the supervised loss function. Hence, we use true supervision where possible (seen classes), and only use pseudo-supervision where there is no alternative (unseen classes). The positive sample for GZSL is therefore chosen as follows
\begin{align}
\be^{+} = \argmin_{\be \in \sE^{s}\cup \sE^{u}} \| \varphi (\sX^{c})-\Theta_2 (\be; W_2) \|_{2}^{2}.\label{arg_GZSL}
\end{align}

The negative sample is selected from the seen semantic embedding set $\sE^{s}$ for both ZSL and GZSL, since all elements of this set will have a different label from the unseen anchor. We choose the negative sample as the seen semantic embedding vector whose projection is closest to the anchor vector $\varphi (\sX^{u})$,
\begin{align}
\be^{-} = \argmin_{\be \in \sE^{s}} \| \varphi (\sX^{s})-\Theta_2 (\be; W_2) \|_{2}^{2}\label{arg_neg}
\end{align}

Finally, the unsupervised loss function $L_{t}$ associated with the unlabeled instances for both ZSL and GZSL tasks is defined as follows:
\begin{align}
    L_{t} =\frac{1}{N^{\prime}}\sum_{i=1}^{N^{\prime}} \max \bigg\{ 0,\left \| \varphi (\sX_{i}^{u})-\Theta_2(\be^{+}; W_2) \right \|_{2}^{2} + m \notag \\ 
    - \left \| \varphi (\sX_{i}^{u})-\Theta_2(\be^{-}; W_2) \right \|_{2}^{2} \bigg\}
    \label{eqn:tripletloss}
\end{align}    
where $m$ is a margin that encourages separation between the clusters, and $N^{\prime}$
is the batch size of the unlabeled instances.

This proposed triplet loss is distinct from recent literature \cite{Facenet,BMVC17Zeroshot} in two ways. (1) Popular methods of triplet formation select a similar feature to the input feature as a positive sample, whereas we choose a semantic word vector for this purpose. This helps to better align the 3D point cloud features with the semantic vectors. (2) We employ a triplet loss in a transductive setting to utilize unlabeled (test) data, whereas established methods consider the triplet loss for inductive training only. This extends the role of the triplet loss beyond inductive learning.

\noindent\textbf{Unsupervised hubness loss:} 
Distance-based ZSL solutions often fall into the trap of the hubness problem. We observe that this issue is intensified for 3D ZSL. To calculate the degree of hubness in a nearest neighbor search problem, the skewness of the empirical distribution $\rho _{j}$ can be used \cite{Shigeto_Hubness_2015,Article57}. The distribution $\rho_{j}$ counts the number of times ($\rho_{j}(i)$) the $i$\textsuperscript{th} point (known as the prototype) is in the top $j$ nearest neighbors of the test samples. The skewness of this distribution is defined as:
\begin{align}
\centering
\rho_{j}\textrm{-skewness}=\frac{\sum_{i=1}^{n}(\rho_{j}(i)-\E\left [ \rho_{j} \right ])^3}{n \left( \Var\left [ \rho_{j} \right ] \right)^{\!\frac{3}{2}}}
\label{eq:skewness_eq}
\end{align}
where $n$ is the number of test prototypes. Large values of skewness indicate that the feature space is severely affected by the hubness problem. In this paper, we mitigate the hubness problem during both inductive and transductive training. As previously discussed, the S2F strategy is effective at reducing hubness during inductive training. We extend this to transductive training by designing a skewness loss based on Eq. \ref{eq:skewness_eq}.

The pseudo-label predicted for the $i$\textsuperscript{th} unlabeled instance of a batch with size $N$ is defined as:
\begin{equation}
\setlength{\jot}{10pt}
    \begin{aligned}
\hat{y}_{i}= \argmin_{\begin{array}{c}\scriptstyle{y\in \mathcal{Y}^{s}\cup \mathcal{Y}^{u}}\\ [-4pt]\scriptstyle {\be \in \sE^{s}\cup \sE^{u}} \end{array}
}\| \varphi (\sX^{c})-\Theta_2 (\be; W_2) \|_{2}^{2}
    \end{aligned}
\end{equation}
Then, for all instances in the batch, we predict their pseudo-labels, and define a set $\hat{\mathcal{T}^{c}}= {\left \{ \hat y_{1},..., \hat y_{N} \right \}}$.
%
%


We calculate the frequency of each class from $\hat{\mathcal{T}}^{c}$ by using the histogram function $\mathcal{H}(\hat{y}_{i})$, which uses counts of the number of times that a specific seen/unseen class is predicted. This function has the property that\\ \noindent$\sum_{i=1}^{S+U}\mathcal{H}(\hat{y}^{c}_{i})=N$. We use the predicted pseudo-labels to find the confidence score. We define the skewness loss as
\begin{align}
L'_{h} = \frac{ 1}{N (\Var[\mathcal{H}(\hat{\mathcal{T}}^{c})])^{\frac{3}{2}}}\sum_{i=1}^{N}(\mathcal{H}(\hat{y}^{c}_{i})-\E[\mathcal{H}(\hat{\mathcal{T}}^{c})])^{3}
\label{eq:skewloss}
\end{align}
where $\mathcal{H}(\hat{\mathcal{T}}^{c})$ represents the statistics of prediction for all instances, that is, how many times each output is predicted regardless of being true or false. The loss $L'_{h}$ tries to balance the number of times a particular class is predicted within a batch and helps the model predict a diverse set of classes. With a larger and randomized batch, the number of particular class instances does not dominate in that batch. As a result, this loss performs better with large batch sizes.

The loss $L'_{h}$ may impact the correct predictions while balancing predicted class distribution. To counter this, inspired by focal loss~\cite{8417976} we weight each sample in the batch based on their confidence in the prediction. To be more specific, if an example in a batch is confident of predicting a pseudo-label, it should contribute less to the hubness loss and vice versa:
\begin{align}
\pi  = -\frac{1}{N}\sum_{i=1}^{N}\log\frac{e^{-\left \|  \varphi (\mathcal{X}^{c}_{i})-\Theta_2 (\be_{\hat y_{i}}; W_2) \right \|^{2}_{2}}}{\sum_{k\in\mathcal{Y}^{s}\cup\mathcal{Y}^{u}}^{}e^{-\left \|  \varphi (\mathcal{X}^{c}_{i})-\Theta_2 (\be_{k}; W_2) \right \|^{2}_{2}}}.
\label{eq:weighthubness}
\end{align}
The final unsupervised hubness loss is given by
\begin{equation}
    L_{h} = \pi L'_{h}.
    \label{eq:finalhubnessloss}
\end{equation}



\noindent\textbf{Unsupervised unbiasing loss:} 
The model has observed many labeled instances during inductive training, and seen class semantics align perfectly with 3D features. As a result, distances between seen semantics and features remain close during transductive learning, but unseen semantics and features reside far apart. This biases the model towards seen classes, confusing unseen instances as seen. Similar to previous work \cite{Song2018TransductiveUE}, we adopt an unsupervised unbiasing loss to minimize this effect.

Given unlabeled examples of the seen and unseen classes, this loss aids our model by increasing unseen probabilities calculated from distances. The model gradually learns to pull unseen semantics close to unseen instances.
\begin{align}
L_{u} = -\frac{1}{N}\sum_{i=1}^{N}\log\!\!\sum_{j\in \mathcal{Y}^{u}}^{}\frac{e^{-\left \|  \varphi (\mathcal{X}^{c}_{i})-\Theta_2 (\be_{j}; W_2)\right \|^{2}_{2}}}{\sum_{k\in\mathcal{Y}^{s}\cup\mathcal{Y}^{u}}^{}e^{-\left \|  \varphi (\mathcal{X}^{c}_{i})-\Theta_2 (\be_{k}; W_2)\right \|^{2}_{2}}}.
\label{eq:unbiasedloss}
\end{align}
This loss balances the average distances between semantic vectors and data features of both seen and unseen classes. Consequently, this process helps the model have less bias towards seen classes, resulting in better accuracy on unseen classes.
 

\noindent\textbf{Overall transductive loss:} The overall loss is given by the sum of the unsupervised triplet, hubness and unbiasing losses as follows:
\begin{equation}
    L_T = \alpha _{1}L_t + \alpha _{2}L_h + \alpha _{3}L_u
\end{equation}
where hyper-parameters $\alpha _{1}$, $\alpha _{2}$, and $\alpha _{3}$ control the importance of $L_{t}$, $L_{h}$, and $L_{u}$ respectively.

\section{Training} 
The proposed model architecture is shown in Figure~\ref{fig:model_architecture}, consisting of two branches: the point cloud network that extracts a feature vector $\varphi(\sX) \in \bbR^{m}$ from a point cloud $\sX$, and the semantic projection network that projects a semantic feature vector $\be\in \bbR^{d}$ into point cloud feature space. Any network that learns a feature space from 3D point sets and is invariant to permutations of points in the point cloud can be used in our method as the point cloud network \cite{Article1,Article2,Article24,Article27,Article28,Article29,Xie_2018_CVPR}. The projection network $\Theta_2(\cdot)$ with trainable weights $W$ consists of two fully-connected layers, with $512$ and $1024$ dimensions respectively, each followed by a $\tanh$ nonlinearity.

In contrast, transductive ZSL additionally uses the set of unlabeled, unseen instances $\{\sX_{i}^{u}\}$ and the set of unseen semantic embedding vectors $\sE^{u}$ during training. To learn a transductive model in a semi-supervised manner, an objective function
\begin{align}
L = L_{S2F} + L_{T}
\label{eqn:LT}
\end{align}
is minimized.

We describe the overall training process in Algorithm \ref{alg:method}. In the proposed algorithm, in the first stage, an inductive model $W_{ind}$ is learned. Then the transductive model $W_{tns}$ is initialized with the inductive model. Finally the transductive model is learned.  

\begin{algorithm}[!t]
\caption{Transductive ZSL for 3D point cloud objects}\label{euclid}
\begin{algorithmic}[1]
\Statex \textbf{Input:} $\sX^{s}$, $\sY^{s}$, $\sE^{s}$, $n_{s}$, $\sX^{u}$, $\sE^{u}$, $n_{u}$
\Statex \textbf{Output:} A trained model $W_{tns}$ to find $\hat{y}$ for all $\sX^{u}$

\Statex \textbf{Inductive training stage}\\
$W_{ind} \gets$ train an inductive model using Eq~\ref{eqn:S2F} with only seen data: $\sX^{s}$, $\sY^{s}$, $\sE^{s}$, $n_{s}$

\Statex \textbf{Transductive training stage}\\
$W_{tns} \gets W_{ind}$, initialize transductive model

\Repeat
\If{GZSL}
\State  $\hat{y} \gets$ use $W_{tns}$ to assign positive and negative anchors to $\sX^{u}$ using Eq~\ref{arg_GZSL} and Eq~\ref{arg_neg} for triple formation
\Else
\State  $\hat{y} \gets$  use $W_{tns}$ to assign  positive and negative anchors to $\sX^{u}$ using Eq~\ref{arg_ZSL} and Eq~\ref{arg_neg} for triple formation
\EndIf

\For{$\forall I \in \sX^{s} \cup   \sX^{u}$}

    \State Calculate triplet loss, $L_t$ using Eq~\ref{eqn:tripletloss}
    \State Calculate hubness loss, $L_h$ using Eq~\ref{eq:skewloss}
    \State Calculate unbiased loss, $L_u$ using Eq~\ref{eq:unbiasedloss}
    \State Calculate overall transductive loss, $L_T$ using Eq~\ref{eqn:LT}
    \State Backpropagate and update $W_{tns}$
\EndFor

\Until convergence

\Statex \textbf{Return} Class decision $\hat{y}$ with $W_{tns}$ using Eq~\ref{eqn:inference_ZSL} for ZSL or Eq~\ref{eqn:inference_GZSL} for GZSL

\end{algorithmic}
\label{alg:method}
\end{algorithm}

\subsection{Inference}
For the zero-shot learning task, given the learned optimal weights $W_2$ from training with labeled seen instances $\sX^{s}$ and unlabeled unseen instances $\sX^{u}$, the label of the input point cloud $\sX^{u}$ is predicted as
\begin{align}
\hat{y} = \argmin_{y \in \sY^{u}} \left\| \varphi(\sX^{u})-\Theta_2(\phi(y); W_2) \right\|_2 \textrm{.} \label{eqn:inference_ZSL}
\end{align}
For the generalized zero-shot learning task, the label of the input point cloud $\sX^{c}$ for $c \in \{s, u\}$ is predicted as
\begin{align}
\hat{y} = \argmin_{y \in \sY^{s} \cup \sY^{u}} \left\| \varphi(\sX^{c})-\Theta_2(\phi(y); W_2) \right\|_2 \textrm{.} \label{eqn:inference_GZSL}
\end{align}

\begin{table}[!t]\centering\small
\newcolumntype{C}{>{\centering\arraybackslash}X}
\renewcommand*{\arraystretch}{1.1}
\setlength{\tabcolsep}{2pt}
\begin{tabularx}{\columnwidth}{l l C C c}\hline
&\multirow{2}{*}{Dataset}    & Total  & Seen/   & Train/ \\ 
& &classes  & Unseen  & Valid/Test \\ \hline
3D&ModelNet40 \cite{Article10} & 40 & 30/--         & 5852/1560/--\\
synt-&ModelNet10 \cite{Article10} & 10 & --/10         & --/--/908\\
hetic&McGill \cite{Article49}     & 19 & --/14         & --/--/115\\ \hline
3D&ModelNet40 \cite{Article10} & 40 & 26/--        & 4999/1496/--\\
real&ScanObjectNN \cite{scanobjectnn_iccv19}  & 15 & /11         & --/--/495\\\hline
\multirow{4}{*}{2D}&AwA2 SS \cite{Xian_CVPR_2017} & 50 & 40/10      &  30337/--/6985\\
&AwA2 PS \cite{Xian_CVPR_2017} & 50 & 40/10      &  23527/5882/7913\\
&CUB SS \cite{CUB_2011} & 200 & 150/50           & 8855/--/2933\\
&CUB PS \cite{CUB_2011} & 200 & 150/50           & 7057/1764/2967\\\hline
\end{tabularx}
\vspace{.2em}
\caption{Statistics of the 3D and 2D datasets. The total number of classes in the datasets are reported, alongside the actual splits used in this paper dividing the classes into seen or unseen and the elements into those used for training or testing. The 3D synthetic splits are from \cite{cheraghian2019zeroshot} and the 2D Standard Splits (SS) and Proposed Splits (PS) are from Xian \textit{et al.} \cite{Xian_CVPR_2017}. The 3D real split is newly proposed in this paper.}
\label{Table:splitting}
\end{table}

\section{Experiments}

\subsection{Setup}

\noindent\textbf{Datasets:} We evaluate our approach on four well-known 3D datasets, ModelNet10~\cite{Article10}, ModelNet40~\cite{Article10}, McGill \cite{Article49}, and ScanObjectNN \cite{scanobjectnn_iccv19}, and two 2D datasets, AwA2 \cite{Xian_CVPR_2017} and CUB~\cite{CUB_2011}. The dataset statistics as used in this work are given in Table \ref{Table:splitting}. We experiment on three different seen/unseen split settings. 
\textbf{\textit{(1)}} For experiments with synthetic datasets (ModelNet10, ModelNet40, and McGill), we follow the seen/unseen splits proposed by Cheraghian et al. \cite{cheraghian2019zeroshot}, where the seen classes are those $30$ in ModelNet40 that do not occur in ModelNet10, and the unseen classes are those from the test sets of ModelNet10 and McGill that are not in the set of seen classes. These splits allow us to test unseen classes from different distributions than that of the seen classes.
\textbf{\textit{(2)}} For experiments with the real 3D dataset (ScanObjectNN), we propose a new train-test setting. Unlike synthetic (CAD) modes of ModelNet40, ScanObjectNN contains real-world scanned objects. We train our method using $26$ non-overlapped classes between ModelNet40 and ScanObjectNN as seen and test with $11$ overlapped classes. This setup uses only ModelNet40 instances during training and ScanObjectNN instances during testing. This is a more realistic setup because we can get many synthetic examples of seen objects during training. However, the model may encounter many real-world 3D data instances of both seen and unseen classes at test time. 
\textbf{\textit{(3)}} For the 2D datasets, we follow the Standard Splits (SS) and Proposed Splits (PS) of Xian et al. ~\cite{Xian_CVPR_2017}.

\begin{figure*}[!t]
\centering
\subfloat[\centering ModelNet40, ModelNet10 and McGill]{{\includegraphics[width=.5\textwidth,trim=1.2cm 2.5cm 0 0, clip]{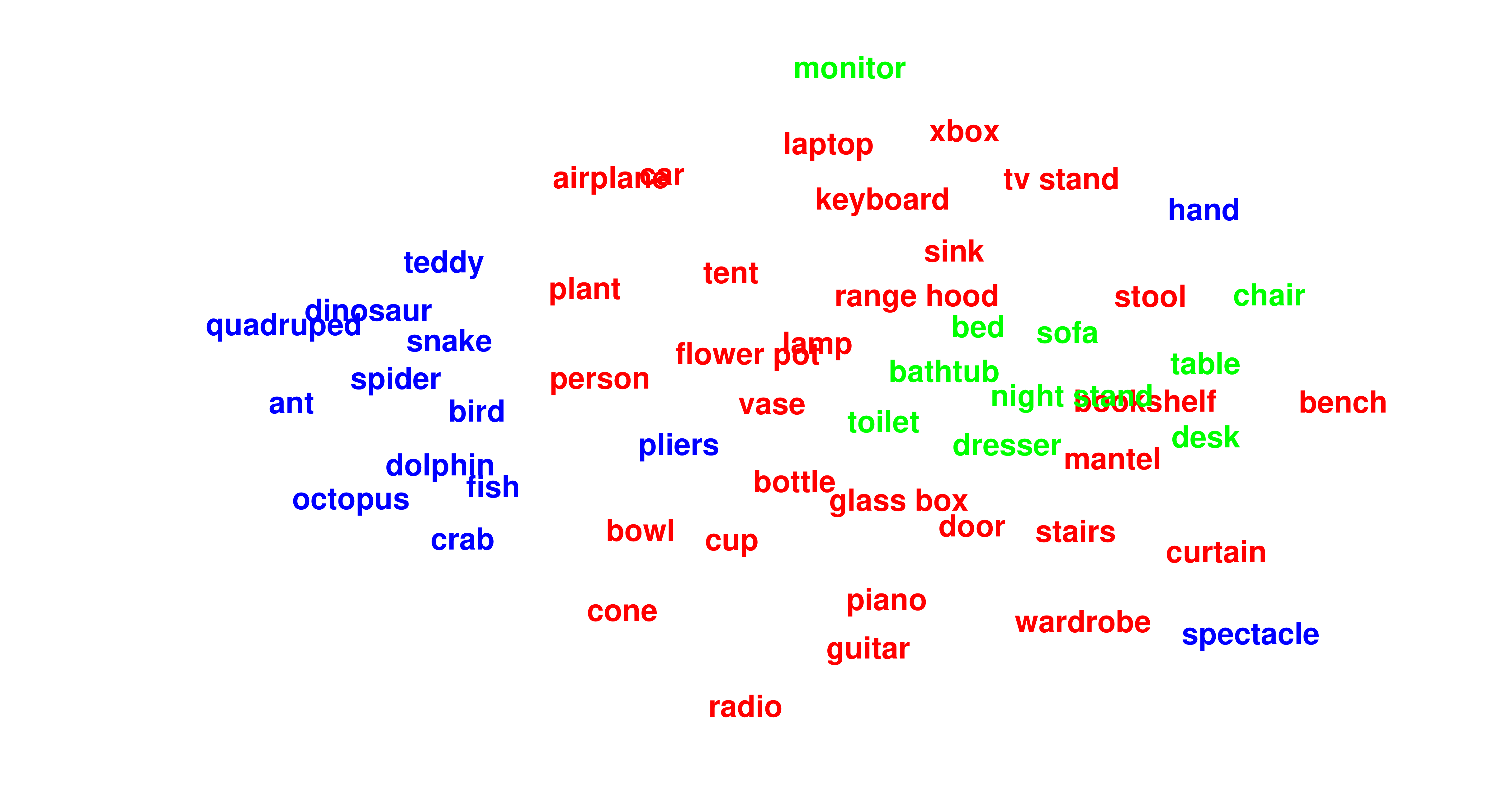} }}%
\qquad
\subfloat[\centering ModelNet40 and ScanObjectNN]{{\includegraphics[width=.4\textwidth, clip]{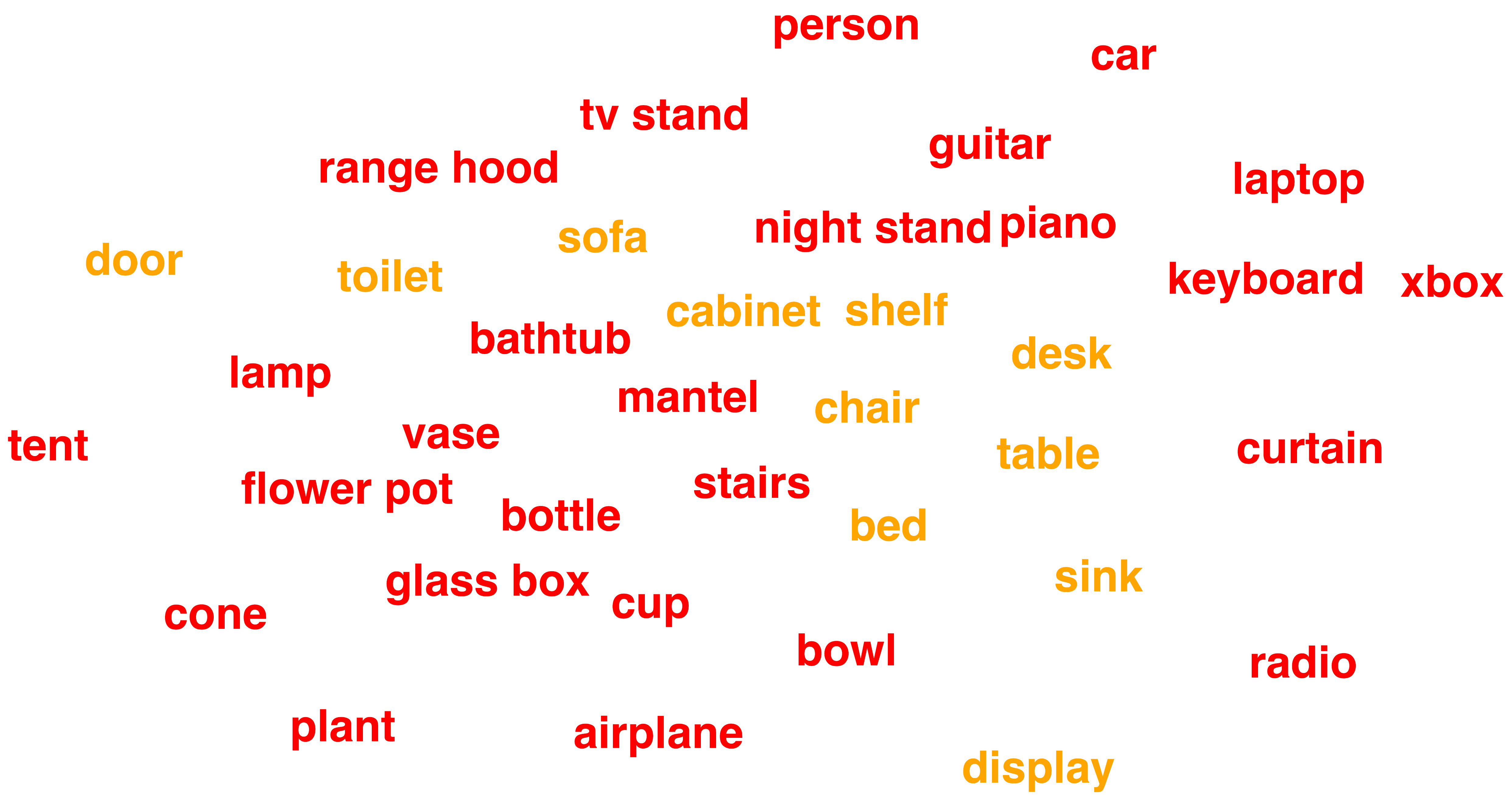} }}%
\caption{2D tSNE~\cite{tSNE_van2014} visualization of word2vec vectors \cite{Mikolov_arXiv_2013}. Red, green, blue and orange texts represent seen {\color{red}ModelNet40 \cite{Article10}}, unseen {\color{green}ModelNet10 \cite{Article10}}, unseen {\color{blue}McGill \cite{Article49}} and unseen {\color{orange}ScanObjectNN \cite{scanobjectnn_iccv19}} classes respectively. 
}

\label{fig:w2vglo}
\end{figure*}

\noindent\textbf{Semantic features:} We use the 300-dimensional semantic feature vectors of word2vec \cite{Mikolov_NIPS_2013}
for the 3D dataset experiments, the 85-dimensional attribute vectors from Xian et al.~\cite{Xian_CVPR_2017} for the AwA2 experiments, and the 312-dimensional attribute vectors from Wah et al.~\cite{CUB_2011} for the CUB experiments. \figref{fig:w2vglo} visualizes word vectors of 3D datasets.

\noindent\textbf{Evaluation:} We report the top-$1$ accuracy as a measure of recognition performance, where the predicted label (the class with minimum distance from the test sample) must match the ground-truth label to be considered a successful prediction. For generalized ZSL, we also report the Harmonic Mean (HM) \cite{Xian_CVPR_2017}  of the accuracy of the seen and unseen classes, computed as
\begin{align}
\textrm{HM} = \frac{2 \times \Acc_{s} \times \Acc_{u} }{\Acc_{s} + \Acc_{u}}
\end{align}
\noindent
where $\Acc_{s}$ and $\Acc_{u}$ are seen and unseen class top-$1$ accuracies respectively. The harmonic mean is able to distinguish between methods that are biased towards seen classes and those that produce good results for both seen and unseen classes.

\noindent\textbf{Cross-validation:} We used cross-validation to find the best hyper-parameters, averaging over $10$ repetitions. For ModelNet10 and McGill, 5 of the 30 seen classes were randomly selected as an unseen validation set, while 4 of the 26 seen classes were chosen randomly for the ScanObjectNN. Additionally, $20\%$ of the seen classes were used as an unseen validation set for the AwA2 and CUB datasets. To find hyperparameters, we conducted a grid search within the range $\alpha_{1}$,$\alpha_{2}$, $\alpha_{3} \in [0,1]$. The selected hyper-parameters $\alpha_1$, $\alpha_2$, and $\alpha_3$ were 0.4, 0.001, 0.001 for both ModelNet10 and McGill, 0.2, 0.2, 0.1 for ScanObjectNN, 0.12, 0.001, 0.01 for AWA, and 0.1, 0.001, 0.001 for CUB.


\noindent\textbf{Implementation details\footnote{Code and data are available at: \url{https://github.com/ali-chr/Transductive_ZSL_3D_Point_Cloud}}:}
For the 3D data experiments, we used PointNet \cite{Article1}, DGCNN\cite{Article24}, PointConv \cite{8954200}, and PointAugment \cite{Li_2020_CVPR} as the point cloud feature extraction network.
For synthetic 3D data, these networks were pre-trained on the 30 seen classes of ModelNet40. Also, for real 3D data, these networks were pre-trained on the 26 seen classes of ModelNet40. For the 2D data experiments, we used a 101-layered ResNet architecture \cite{He2016DeepRL}, where the 2048-dimensional input feature embedding was obtained from the top-layer pooling unit. The network was pre-trained on ImageNet 1K \cite{imagenet_cvpr09}. 
For semantic projection layers, we used two fully connected  (512,1024) with relu non-linearities for 3D experiments, and two fully connected  (1024,2048) with relu non-lin\-earities for 2D experiments. These parameters are fully-learnable.
To train the network, we used the Adam optimizer~\cite{Article40} with an initial learning rate of 0.0001 
for all experiments. We implemented the architecture using Pytorch and trained and tested it on a NVIDIA GTX Titan~V GPU.

\noindent\textbf{Compared approaches:} We enlist different versions (baselines and our recommendation) of the proposed method below:

\begin{itemize}\setlength{\itemsep}{-0.2em}

\item \textit{Baseline-I:} Our inductive baseline while projecting point cloud feature to semantic embedding space (F2S) using Eq. \ref{eqn:F2S}.

\item \textit{Ours (Inductive):} Our recommended inductive approach while projecting semantic embedding space to point cloud feature (S2F) using Eq. \ref{eqn:S2F}

\item \textit{Baseline-T:} Our transductive baseline using only triplet loss of Eq. \ref{eqn:tripletloss} as transductive loss (without hubness and unbiased loss part), i.e., $L_T = L_{t}$.

\item \textit{Ours (Transductive):} Our recommended transductive approach while using Eq. \ref{eqn:LT}.


\end{itemize}




\begin{table}[!t]\centering \small
\newcolumntype{C}{>{\centering\arraybackslash}X}
\renewcommand*{\arraystretch}{1.1}
\setlength{\tabcolsep}{2pt}
\begin{tabularx}\columnwidth{lCCC}\hline
Backbone & All-40 & Seen-30 & Seen-26\\\hline
PointNet~\cite{Article1} & 89.2 & 85.7 & 87.1 \\ 
PointAugment~\cite{Li_2020_CVPR} & 90.9 & 88.3 & 89.5 \\ 
DGCNN~\cite{Article24} & \textbf{92.2} & 91.2 & 92.5 \\ 
PointConv~\cite{8954200} & \textbf{92.2} & \textbf{92.6} & \textbf{93.1} \\ 
\hline
\end{tabularx}
\caption{Results on seen classes of ModelNet40 for different feature extractor backbones.}
\label{Table:SeenClasses}
\end{table}

\begin{table*}[!t]
\centering \small
\newcolumntype{C}{>{\centering\arraybackslash}X}
\setlength{\tabcolsep}{2pt}
\begin{tabularx}\textwidth{ccCCCC|CCCC}\hline
\multirow{3}{*}{Backbone} & \multirow{3}{*}{Method} & \multicolumn{4}{c}{ModelNet10} & \multicolumn{4}{c}{ScanObjectNN} \\ 
\cline{3-10}
{} & & ZSL & \multicolumn{3}{c}{GZSL} & ZSL & \multicolumn{3}{c}{GZSL}\\ \cline{3-10}
& & Acc & $\Acc_{s}$ & $\Acc_{u}$ & HM & Acc & $\Acc_{s}$ & $\Acc_{u}$ & HM \\ \hline

\multirow{2}{*}{PointNet} & Ours (Inductive) & 21.26 & 79.37 & 3.74 & 7.15 & 18.95 & 75.13 & 3.58 & 6.83\\
 & Ours (Transductive) & 18.28 & 71.79 & 16.08 & 26.27 & 18.11 & 75.27 & 6.11 & 11.29 \\
 \hline
 
\multirow{2}{*}{PointAugment} & Ours (Inductive) & 21.37 & 71.92 & 6.39 & 11.73 & 16.42 & 52.14 & 2.74 & 5.20 \\
 & Ours (Transductive) & 23.68 & 66.86 & 12.67 & 21.30 & 18.95 & 40.64 & 14.32 & 21.17 \\
 \hline
 
\multirow{2}{*}{DGCNN} & Ours (Inductive) & 38.33 & 69.87 & 8.26 & 14.77 & 22.95 & 79.28 & 1.89 & 3.70 \\
 & Ours (Transductive) & {60.05} & {78.71} & {45.26} & {57.47} & 25.68 & 53.54 & 9.89 & 16.70 \\
 \hline
 
 \multirow{2}{*}{PointConv}& Ours (Inductive) & 32.49 & \textbf{89.42} & 6.83 & 12.69 & 21.89 & 89.37 & 5.68 & 10.69 \\
 & Ours (Transductive) & \textbf{68.50} & 83.21 & \textbf{65.64} & \textbf{73.39} & \textbf{30.53} & \textbf{90.31} & \textbf{30.53} & \textbf{45.63}\\
 \hline
\end{tabularx}
\caption{Performance of our method using different backbones.} 
\label{tab:allbackbone}
\end{table*}

\begin{table*}[!t] \centering \small
\newcolumntype{C}{>{\centering\arraybackslash}X}
\setlength{\tabcolsep}{4pt}

\begin{tabularx}{\textwidth}{clCCCC|CCCC|CCCC}\hline
\multirow{3}{*}{} & \multirow{3}{*}{Method (PointConv)} & \multicolumn{4}{c}{ModelNet10} & \multicolumn{4}{c}{McGill} & \multicolumn{4}{c}{ScanObjectNN} \\
\cline{3-14}
{} & {} & ZSL & \multicolumn{3}{c}{GZSL} & ZSL & \multicolumn{3}{c}{GZSL} & ZSL & \multicolumn{3}{c}{GZSL} \\
\cline{3-14}
 &  & Acc & $\Acc_{s}$ & $\Acc_{u}$ & HM & Acc & $\Acc_{s}$ & $\Acc_{u}$ & HM & Acc & $\Acc_{s}$ & $\Acc_{u}$ & HM \\ \hline
\multirow{9}{*}{I}  
& DEM \cite{dem-cvpr17} & 17.48 & 88.57 & 5.30 & 9.99 & 7.12 & 75.95 & 7.14 & 13.06 & 10.71 & 88.76 & 10.71 & 19.12\\  
& LATEM \cite{latem-cvpr16} & 26.29 & - & - & - & 7.15 & - & - & - & 11.88 & - & - & -\\ 
& SYNC \cite{sync-cvpr16} & 21.17 & - & - & - & 7.14 & - & - & - & 17.43 & - & -& -\\ 
& GDAN \cite{gdan-cvpr19} & - & 86.57 & 4.06 & 7.76 & - & 86.97 & 7.14 & 13.20 & - & 88.34 & 19.07 & 31.37\\ 
& TF-VAEGAN \cite{tfvaegan-eccv20} & 27.21 & 59.23 & 19.65 & 29.51 & \textbf{20.65} & 84.63 & \textbf{20.65} & \textbf{33.19} & 28.20 & 81.22 & 23.99 & 37.04\\
& f-CLSWGAN  \cite{Xian_2018_CVPR} & 13.73 & 67.13 & 15.57 & 25.27 & 17.21 & 85.13 & 13.57 & 23.41 & 18.35 & 85.60 & 11.61 & 20.44\\
& CADA-VAE \cite{Schonfeld_2019_CVPR} & 15.58 & 89.1 & 2.93 & 5.67 & 7.14 & 89.27 & 7.14 & 13.23 & 16.47 & 89.61 & 14.11 & 24.38\\
& Baseline-I & 24.45 & 27.12 & 8.81 & 13.30 & 13.04 & 62.69 & 0.00 & 0.00 & 21.68 & 37.10 & 1.05 & 2.05\\
& Ours (Inductive) & 32.49 & \textbf{89.42} & 6.83 & 12.69 & 13.91 & 90.51 & 13.91 & 14.39 & 24.12 & 89.37 & 5.68 & 10.69\\
\hline
\multirow{3}{*}{T} & QFSL \cite{Song2018TransductiveUE} & 38.80 & 58.10 & 21.80 & 31.70 & 9.56 & 86.08 & 9.56 & 17.21 & 18.71 & 81.88 & 18.53 & 30.21 \\
& Baseline-T & 43.17 & 85.58 & 42.96 & 57.20 & 5.22 & \textbf{90.71} & 5.22 & 9.87 & 25.05 & 88.50 & 25.05 & 39.05\\
& Ours (Transductive) & \textbf{68.50} & 83.21 & \textbf{65.64} & \textbf{73.39} & 15.71 & 71.08 & 8.69 & 15.49 & \textbf{30.53} & \textbf{90.31} & \textbf{30.53} & \textbf{45.63}\\ \hline
\end{tabularx}
\vspace{.2em}
\caption{ ZSL and GZSL results on the 3D ModelNet10 \cite{Article10}, McGill \cite{Article49}, and ScanObjectNN \cite{scanobjectnn_iccv19} datasets for 
PointConv~\cite{8954200}. We report the top-1 accuracy (\%) on seen classes ($\Acc_s$) and unseen classes ($\Acc_u$) for each method, as well as the harmonic mean (HM) of both measures. ``I'' and ``T'' denote inductive and transductive learning respectively.}
\label{table:GZSL_3D}
\end{table*}




\subsection{Comparing point cloud feature extractors}

\begin{figure*}
\centering
\includegraphics[width=1\linewidth,trim=0cm 0cm 0cm 0cm, clip]{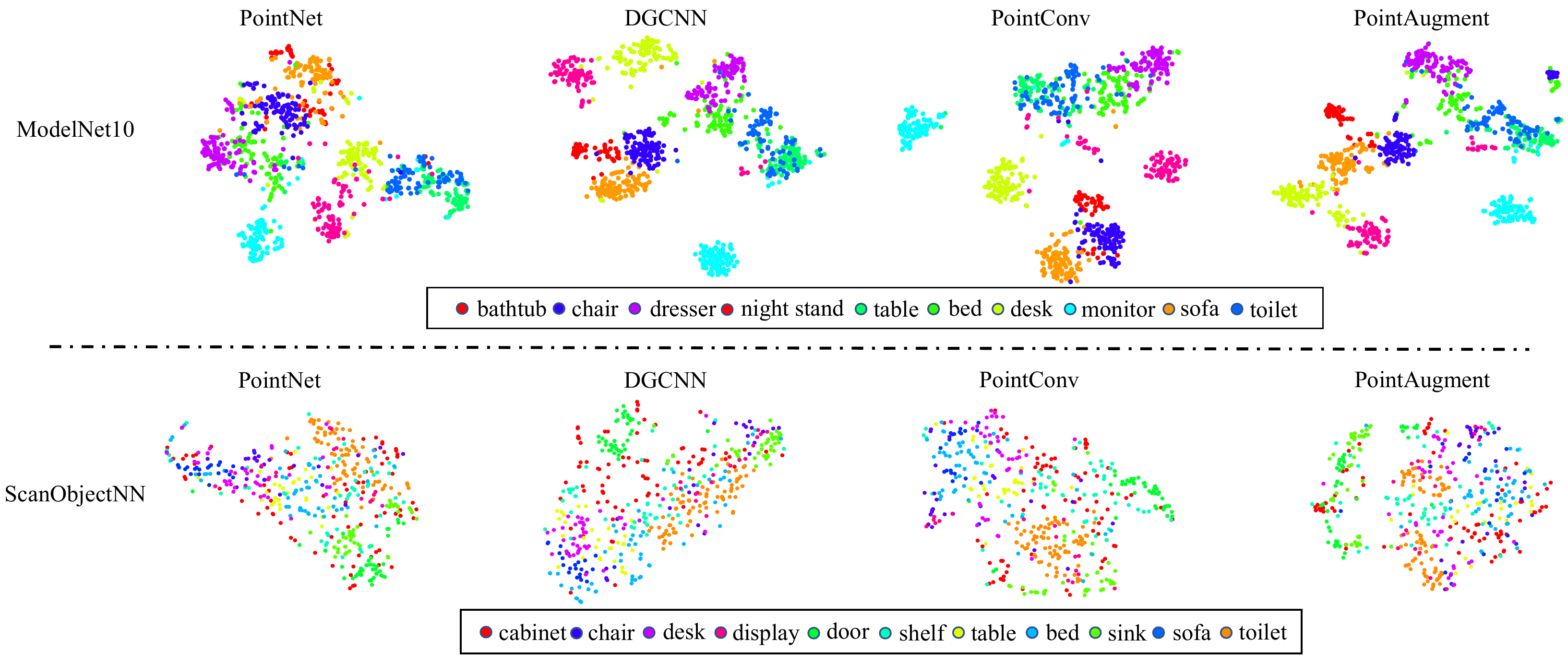}
\caption{2D tSNE~\cite{tSNE_van2014} visualization of unseen point cloud feature vectors (circles) with four backbone networks. Since the synthetic data has less noise, ModelNet40 features are clustered better than the real scanned 3D data (with noise) from ScanObjectNN.
Moreover, for both datasets, the models are trained on synthetic instances belonging to a subset of ModelNet40 classes, and so we expect the ModelNet10 features to be better clustered than the ScanObjectNN features.
We obtained the best overall performance using the PointConv backbone.
}
\label{fig:point_cloud_architecture}
\end{figure*}

We evaluate four 3D point cloud recognition frameworks, namely, PointNet \cite{Article1},  DGCNN \cite{Article24}, PointConv \cite{8954200}, and PointAugment \cite{Li_2020_CVPR} as backbone to extract 3D point cloud features. In Figure \ref{fig:point_cloud_architecture}, we visualize point cloud features for unseen ModelNet10 and ScanObjectNN instances using tSNE. We perform the inductive training (S2F) of all those frameworks on 30 and 26 seen classes of the synthetic ModelNet40 dataset. In Table \ref{Table:SeenClasses}, we report the performance of test seen classes.
The values with 40 classes (All-40) of ModelNet40 are from the original published papers. The columns for Seen-30 and Seen-26 report the performance of 30 and 26 seen classes during training with synthetic and real-world scanned 3D datasets, respectively. We notice similar performance for All-40, Seen-30, and Seen-26 experiment setups, which tells that the backbone is well-trained for feature extraction.


In addition to test seen class performance, in Table \ref{tab:allbackbone} we show ZSL and GZSL results of the same inductive training using test samples from seen and unseen classes from both synthetic and real 3D datasets. From Table \ref{Table:SeenClasses} and \ref{tab:allbackbone}, we notice that DGCNN and PointConv point cloud backbone performs consistently better than other. The reason is that DGCNN and PointConv analyze the local and global information of point cloud data, while PointNet and PointAugment consider solely global information. 
We choose the best performing backbone, PointConv, for the remaining experiments in this paper. 

\begin{figure}[!t]
\centering
\includegraphics[width=.5\textwidth, trim={0.25cm 0 0 0}, clip]{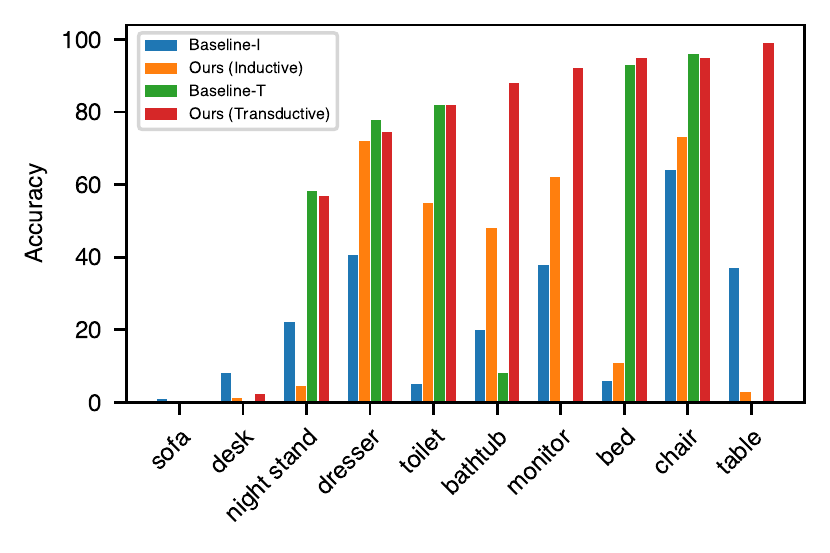}
\caption{ZSL per-class accuracy for ModelNet10 using PointConv backbone.} 

\label{fig:per_class_accuracy}

\end{figure}

\subsection{3D point cloud experiments}

For the experiments on 3D data, we compare different (baselines and recommended) versions  of our method with 
eight 2D ZSL methods, DEM \cite{dem-cvpr17}, SYNC \cite{sync-cvpr16}, LA\-TEM \cite{latem-cvpr16}, GDAN \cite{gdan-cvpr19}, TF-VAEGAN \cite{tfvaegan-eccv20}, f-CLSWGAN \cite{Xian_2018_CVPR}, CADA-VAE \cite{Schonfeld_2019_CVPR}, and QFSL \cite{Song2018TransductiveUE} in Table \ref{table:GZSL_3D}. These state-of-the-art image-based methods were re-implem\-ented and adapted to point cloud data to facilitate comparison. Our method significantly outperforms the other approaches on the ModelNet10 and ScanObjectNN datasets. Several observations can be made from the results.
\textbf{(1)} Methods usually work better on the 3D synthetic dataset (ModelNet10) than real data (ScanObjectNN). This is likely due to domain shift from synthetic to real data and the presence of noise in real data. 
(see \figref{fig:point_cloud_architecture}). However, methods do not perform as well on the McGill dataset when compared to the ModelNet10 results, because the distributions of semantic feature vectors in the unseen McGill datasets are significantly different from the distribution in the seen ModelNet40 dataset, much more so than that of ModelNet10  (see Figure \ref{fig:w2vglo}).
\textbf{(2)} 2D ZSL methods can perform 3D ZSL using 3D features as input instead of 2D images. Generative methods (TF-VAEGAN, CADA-VAE) perform better than non-generative methods (DEM, SYNC) because generative models use unseen semantics during training to create fake features. 
\textbf{(3)} Transductive learning is much more effective than inductive learning for point cloud ZSL. This is likely due to inductive approaches being more biased towards seen classes, while transductive approaches alleviate the bias problem by using unlabeled, unseen instances during training.
\textbf{(4)} Our proposed method performs better than QFSL, which is likely due to our triplet loss formulation. While noisy, the positive and negative samples of unlabeled data provide useful supervision, unlike the unsupervised approach for only unlabeled data in QFSL.
\textbf{(5)} There is a performance improvement from Baseline-T to Ours (Transductive) due to the use of the hubness and unbiasing losses, which mostly contribute to improving ZSL and GZSL performances, respectively.
\textbf{(6)} Our method could not achieve the best performance on McGill because of fewer test instances (more specifically, only 115 instances) available for this dataset (see \tabref{Table:splitting}). However, methods like TF-VAEGAN \cite{tfvaegan-eccv20} and f-CLSWGAN  \cite{Xian_2018_CVPR}) are relatively successful because of generating pseudo-features with generative models, which balances the number of unseen instances similar to seen class instances.
\textbf{(7)} Generalized ZSL, which is more realistic than standard ZSL, is more challenging than ZSL as there are both seen and unseen classes during inference. Our (Transductive) method obtained the best performance with respect to the harmonic mean (HM) on all datasets (not on McGill), and the best performance with respect to the unseen class accuracy $\Acc_{u}$ on most datasets, which demonstrates the utility of our method for GZSL as well as ZSL for 3D point cloud recognition.

\noindent\textbf{Per-class results:} We also show, in \figref{fig:per_class_accuracy}, the performance of individual classes from ModelNet10. Baseline-I performs relatively well (above 30\%) on only four classes (dresser, monitor, chair and table) of ModelNet40. Because of the hubness problem, Baseline-I mostly predicts those few classes regardless of the input. Our inductive and Baseline-T methods minimize this problem by confidently predicting more (five) classes than Baseline-I. Our final transductive method achieves the best accuracy in eight classes and outperforms its alternatives. This is likely due to minimizing the hubness and bias problem in transductive settings.


\begin{table*}[!t]
\centering \small
\newcolumntype{C}{>{\centering\arraybackslash}X}
\setlength{\tabcolsep}{4pt}
\begin{tabularx}\textwidth{CCCCCCCCCC}\hline
\multirow{2}{*}{Backbone} & \multirow{2}{*}{F2S} & \multirow{2}{*}{S2F} & \multirow{2}{*}{Triplet} & \multirow{2}{*}{Hubness} & \multirow{2}{*}{Unbiasing} & \multicolumn{2}{c}{ModelNet10} & \multicolumn{2}{c}{ScanObjectNN} \\ 
\cline{7-10} 
& & & & & & \multicolumn{1}{c}{ZSL} & \multicolumn{1}{c}{GZSL (HM)} & \multicolumn{1}{c}{ZSL} & \multicolumn{1}{c}{GZSL (HM)} \\ \hline

\multirow{5}{*}{PointConv} & \cmark& \xmark & \xmark & \xmark & \xmark & 24.45 & 13.30 & 21.68 & 2.05\\
 & \xmark& \cmark & \xmark & \xmark & \xmark & 32.49 & 12.69 & 21.89 & 10.69 \\
 & \xmark& \cmark & \cmark & \xmark & \xmark & 43.17 & 34.83 & 25.05 & 20.65 \\
 & \xmark& \cmark & \cmark & \cmark & \xmark & 61.45 & 57.33 & 30.74 & 33.59 \\
 & \xmark& \cmark & \cmark & \cmark & \cmark & \textbf{68.50} & \textbf{73.39} & \textbf{30.53} & \textbf{45.63} \\
 \hline

\end{tabularx}
\caption{Ablation studies. Effect of adding different loss components incrementally.
} 
\label{tab:ablationacc}
\end{table*}

\begin{figure}[!t]
\centering
\includegraphics[trim={0.2cm .2cm 0 0},clip]{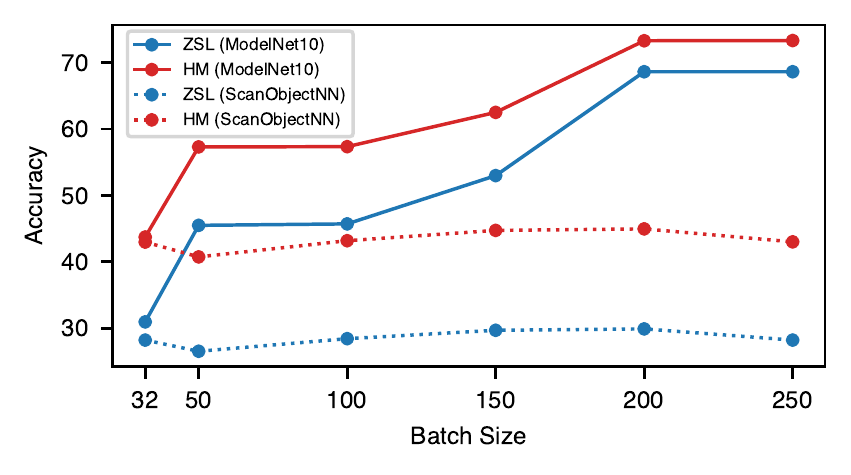}
\vspace{-2em}
\caption{Ours (Transductive) performance when varying batch sizes for ModelNet10 and ScanObjectNN.} 
\label{fig:batch}
\end{figure}

\begin{table}[!t]
\centering \small
\newcolumntype{C}{>{\centering\arraybackslash}X}
\setlength{\tabcolsep}{4pt}
\begin{tabularx}\linewidth{cc|cccc}\hline
\multicolumn{2}{c|}{} & \multicolumn{2}{c}{ModelNet10} & \multicolumn{2}{c}{ScanObjectNN}\\
\cline{3-6}
\multicolumn{2}{c|}{} & \multicolumn{1}{p{1.3cm}}{\multirow{2}{*}{$\pi=1$}} & \multicolumn{1}{p{1.3cm}}{\centering $\pi$ as in \\ Eq. \ref{eq:weighthubness}} & \multicolumn{1}{p{1.3cm}}{\multirow{2}{*}{$\pi=1$}} & \multicolumn{1}{p{1.3cm}}{\centering $\pi$ as in \\ Eq. \ref{eq:weighthubness}}\\
\hline
ZSL & Acc & 46.37 & 68.50 & 19.58 & 30.53 \\
\hline
\multirow{3}{*}{GZSL} & $\Acc_{s}$ & 80.06 & 83.21 & 76.87 & 90.31 \\
& $\Acc_{u}$ & 44.27 & 65.64 & 17.26 & 30.53\\
& HM & 57.02 & 73.39 & 28.19 & 45.63\\
\hline
\end{tabularx}
\caption{The effect of weighting factor, $\pi$ in Eq. \ref{eq:weighthubness} on Ours (Transductive) performance.}
\label{tab:ablation_weight}
\vspace{4mm}
\end{table}

\begin{table*}[!t] \centering \small
\newcolumntype{C}{>{\centering\arraybackslash}X}
\setlength{\tabcolsep}{7pt}
\begin{tabularx}{\textwidth}{clCCCCC|CCCCC}
\hline
\multirow{3}{*}{} & \multirow{3}{*}{Method} & \multicolumn{5}{c}{AwA2} & \multicolumn{5}{c}{CUB} \\
\cline{3-12}
{} & {} & \multicolumn{2}{c}{ZSL} & \multicolumn{3}{c}{GZSL} & \multicolumn{2}{c}{ZSL} & \multicolumn{3}{c}{GZSL}\\
\cline{3-12}
 &  & SS & PS & $\Acc_{s}$ & $\Acc_{u}$ & HM & SS & PS & $\Acc_{s}$ & $\Acc_{u}$ & HM \\ \hline
\multirow{12}{*}{I} & SJE \cite{Akata_CVPR_2015} & 69.5 & 61.9 & - & - & - & 55.3 & 53.9 & - & - & -\\
& DEM\cite{dem-cvpr17} & - & 67.1 & 30.5 & 86.4 & 45.1 & 58.3 & 51.7 & 19.6 & 57.9 & 29.2  \\
& CS\cite{Chao_ECCV_2016} & - & 77.6 & 45.3 & 57.2 & - & 49.4 & 48.1 & 48.7 \\ 
& TCN\cite{Jiang_2019_ICCV} & - & 71.2 & 65.8 & 61.2 &63.4 &  -& 59.5 & 52.0  & 52.6  & 52.3 \\
& GDAN\cite{gdan-cvpr19} & - & - & 67.5 & 32.1 & 43.5 & - & - & 75.0 & 30.4 & 43.4 \\
& TF-VAEGAN\cite{tfvaegan-eccv20} & - & 72.2 & 75.1 & 59.8 & 66.6 & - & 64.9 & 64.7 & 52.8 & 58.1 \\
& TF-VAEGAN*\cite{tfvaegan-eccv20} & - & 73.4 & 83.6 & 55.5 & 66.7 & - & 74.3 & 79.3 & 63.8 & 70.7\\
& f-CLSWGAN  \cite{Xian_2018_CVPR} & - & - & - & - & - & - & 57.3 & 43.7 & 57.7 & 49.7\\
& CADA-VAE \cite{Schonfeld_2019_CVPR} & - & - & 75.0  & 55.8 & 63.9 & - & - & 53.5 & 51.6 & 52.6 \\
& f-VAEGAN-D2 \cite{Xian_2019_CVPR} & - & 71.1 & 57.6 & 70.6 & 63.5 & - & 72.9 & 48.4 & 60.1 & 53.6 \\
& f-VAEGAN-D2{*} \cite{Xian_2019_CVPR} & - & 70.3 & 57.1 & 76.1 & 65.2 & - & 72.9 & 63.2 & 75.6 & 68.9\\
& Ours (Inductive) & 71.2 & 69.0 & 88.9 & 22.1 & 35.4 & 59.3 & 54.2 & 69.4 & 8.4 & 14.9\\ \hline

\multirow{11}{*}{T} & DIPL\cite{Zhao_NIPS_2018} & - & - & - & - & - & 68.2 & 65.4 & 44.8 & 41.7 & 43.2 \\
& PREN \cite{pren} & - & 74.1 & 88.6 & 32.4 & 47.4 & - & 66.4 & 55.8 & 35.2 & 43.1\\
& EDE\_ex \cite{ede_ex} & - & 77.5 & 93.2 & 68.4 & 78.9 & - & 67.8 & 62.9 & 54.0 & 58.1\\
& QFSL{*}\cite{Song2018TransductiveUE} & 84.8 & 79.7 & \textbf{93.1} & 66.2 & 77.4 & 69.7 & 72.1 & \textbf{74.9} & 71.5 & 73.2 \\
& GMN \cite{gmn} & - & - & - & - & - & - & 64.6 & 70.6 & 60.2 & 65.0\\
& f-VAEGAN-D2 \cite{Xian_2019_CVPR} & - & {89.8} & 84.8 & 88.6 & 86.7 & - & 71.1 & 61.4 & 65.4 & 63.2\\
& f-VAEGAN-D2{*} \cite{Xian_2019_CVPR} & - & 89.3 & 86.3 & \textbf{88.7} & \textbf{87.5} & - & \textbf{82.6} & 73.8 &\textbf{81.4}  &\textbf{77.3} \\
& TF-VAEGAN\cite{tfvaegan-eccv20} & - & 92.1 & 89.6 & 87.3 & 88.4 & - & 74.7 & 72.1 & 69.9 & 71.0\\
& TF-VAEGAN*\cite{tfvaegan-eccv20} & - & 93.0 & 90.0 & 89.2 & 89.6 & - & 85.1 & 83.5 & 78.4 & 80.9\\
& Baseline-T & 83.3 & 75.6 & 88.0 & 67.2 & 76.2 & 70.6 & 58.3 & 51.4 & 40.2 & 45.1 \\
& Ours (Transductive) & \textbf{91.2} & \textbf{90.2} & 84.7 & {81.9} & {83.3} & \textbf{72.0} & 71.5 & 60.2 & 58.7 & 59.8 \\ \hline
\end{tabularx}
\vspace{.2em}
\caption{ZSL results on the Standard Splits (SS) and Proposed Splits (PS) and GZSL results on the 2D AwA2 and CUB datasets. We report the top-1 accuracy (\%) on seen classes ($\Acc_s$) and unseen classes ($\Acc_u$) for each method, as well as the harmonic mean (HM) of both measures. ``I'' and ``T'' denote inductive and transductive learning respectively. $^{*}$Image feature extraction model fine-tuned (we do not fine-tune our model).}
\label{table:GZSL_2D}
\vspace{-.2em}
\end{table*}

\subsection{Ablation Studies}

\noindent\textbf{Impact of loss components:} We ablate our proposed method with respect to the different loss components. The elements of the combined loss function incrementally bring robustness to our approach. Table \ref{tab:ablationacc} reports the ablation results with PointConv backbone. 
Our method performs poorly with only the F2S part (Eq. \ref{eqn:F2S}), largely because of the hubness problem. Replacing F2S with S2F (Eq. \ref{eqn:S2F}) improves performance since this mitigates the hubness problem for inductive learning. We perform transductive training based on unsupervised triplet loss (Eq. \ref{eqn:tripletloss}) on top of the S2F-based inductive weights. The utilization of unlabeled data raises the performance from inductive to transductive settings. Next, we add the hubness loss (Eq. \ref{eq:finalhubnessloss}) to minimize the hubness problem further in transductive settings.
Finally, we include the unbiasing loss (Eq. \ref{eq:unbiasedloss}) to balance seen and unseen class distances. It mostly helps to achieve robust GZSL performance because GZSL considers both seen and unseen classes together. 


\noindent\textbf{Impact of batch size:} Our proposed transductive loss has a noticeable impact on the batch size.  With a larger batch size, the mistakes of pseudo-labeling while calculating the triplet loss (Eq. \ref{eqn:tripletloss}) become stabilized. Moreover, it also increases the chance of evenly distributing different class instances, which estimates the hubness loss of Eq. \ref{eq:finalhubnessloss} better than small-batch cases. In Figure~\ref{fig:batch}, we report ZSL and GZSL (HM) performance on ModelNet10 and ScanObjectNN using different batch sizes. As expected, increasing batch size improves the performance.

\noindent\textbf{Impact of weighting factor of hubness loss:} In Eq. \ref{eq:weighthubness}, we design a weight factor for the hubness loss to penalize highly confident predictions less than low scores. In \tabref{tab:ablation_weight}, we show the impact of using the weighting factor. We notice that the use of the weighting factor significantly improves the performance.

\begin{figure*}[!t]
\centering
\includegraphics[width=1\textwidth]{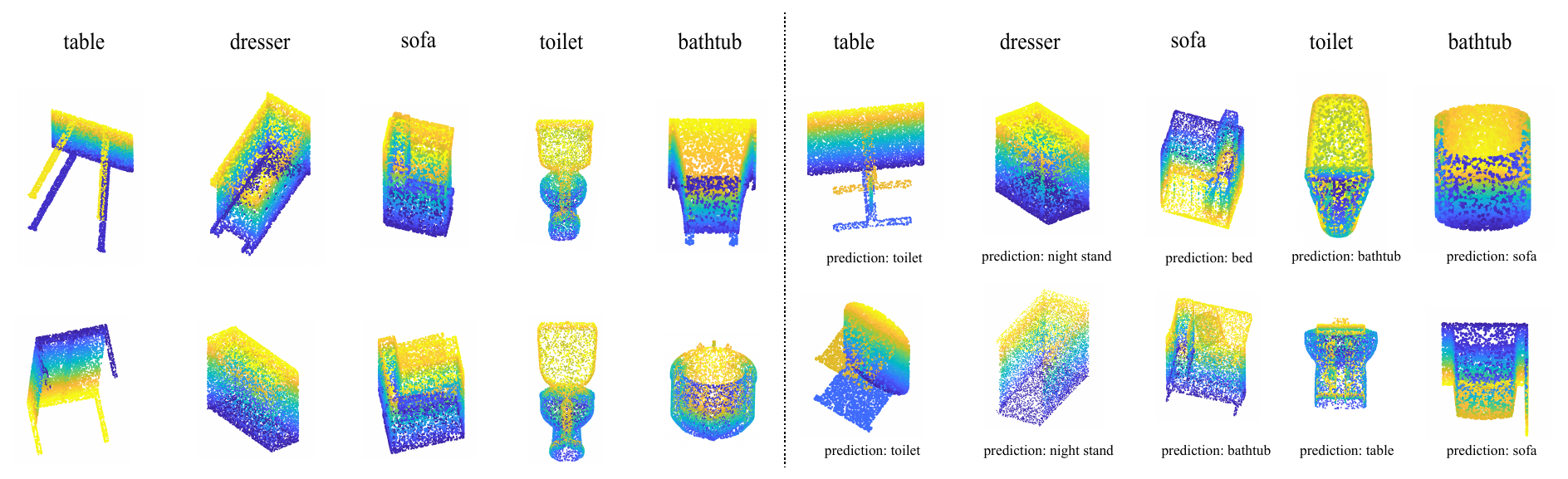}
\caption{Visualization of five classes from the ModelNet10 dataset with examples of (left) correctly and (right) incorrectly classified point clouds, respectively. The predicted classes are shown below each model for the incorrect cases.}
\label{fig:failure}
\end{figure*}

\subsection{2D Image Experiments}

While our method is designed to address ZSL and GZSL tasks for 3D point cloud recognition, we also adapt and evaluate our method for the case of 2D image recognition. The results for ZSL and GZSL are shown in Table \ref{table:GZSL_2D}. Our proposed method is evaluated on the AwA2~\cite{Xian_CVPR_2017} and CUB~\cite{CUB_2011} datasets using the SS and PS splits \cite{Xian_CVPR_2017}. We achieve very competitive results on these datasets, indicating that the method can generalize to image data. Although we outperform many state-of-the-art methods in transductive ZSL settings,  our results lag state-of-the-art in the GZSL problem. Fine-tuning the feature extraction network~\cite{Song2018TransductiveUE} or pseudo-features from generative models~\cite{Xian_2018_CVPR,Schonfeld_2019_CVPR,Xian_2019_CVPR,tfvaegan-eccv20} may go some way to closing this gap.

Another observation, the 2D image experiment results are better in general than 3D point cloud experiments in \tabref{table:GZSL_3D}. The possible reasons could be the availability of large-scale 2D datasets, pre-trained models, and more accurate pseudo/fake features from generative models compared to that of 3D point cloud objects.

\subsection{Qualitative Evaluation}

We visualize five unseen classes from the ModelNet10 dataset with examples where our method correctly classified the point cloud, shown in Figure \ref{fig:failure} (left), and examples where it incorrectly classified the point cloud, shown in Figure \ref{fig:failure} (right). The network appears to be providing incorrect predictions for mostly hard examples, those that are quite different from standard examples in that class, or where the classes overlap in their geometry, such as dresser and night stand.


\subsection{Discussion}

\noindent\textbf{Challenges with 3D data:} Recent deep learning methods for classifying point cloud objects have achieved over 90\% accuracy on several standard datasets, including ModelNet40 and ModelNet10. Moreover, due to significant progress in depth camera technology ~\cite{rs10020328,Izadi_3D_2011}, it is now possible to capture 3D point cloud objects at scale much more easily. It is therefore likely that many classes of 3D objects will not be present in the labeled training set. As a result, zero-shot classification systems will be needed to leverage other more easily-obtainable sources of information in order to classify unseen objects. However, we observe that the difference in accuracy between ZSL and supervised learning is still very large for 3D point cloud classification, e.g., 68.5\% as compared to 95.7\% \cite{Article27} for ModelNet10. As such, there is significant potential for improvement for zero-shot 3D point cloud classification. While the performance is still quite low, this is also the case for 2D ZSL, with state-of-the-art being 31.1\% top-5 accuracy on the ImageNet2010/12~\cite{ILSVRC_2015} datasets, reflecting the challenging nature of the problem.

\noindent\textbf{Visual features versus point cloud features:} Moving from 2D visual features to 3D point cloud features for ZSL brings new challenges. Many deep learning models on 2D images rely on pre-trained deep features, which are obtained by considering thousands of classes and millions of images~\cite{Xian_CVPR_2017}. In contrast, 3D point cloud datasets of a similar scale are not yet available. Therefore, 3D point cloud features are less robust than image features. To illustrate this point, we visualize 6985 instances of 10 classes from the 2D dataset AwA2~\cite{Xian_CVPR_2017} and 908 instances of 10 classes from the 3D dataset ModelNet10~\cite{Article10} in \figref{fig:2D_versus_3D}. It is apparent that although a larger number of instances were used in the 2D case, the cluster structure is more separable in 2D than in 3D. As 3D features are not as robust and separable as 2D features, relating those features to their corresponding semantic vectors is more difficult in 3D than 2D.

\noindent\textbf{Hubness:} 
Our approach, projecting semantic vectors to input feature (S2F) space, since it has been shown that this alleviates the hubness problem \cite{Shigeto_Hubness_2015,Zhang_2017_CVPR}, we validate this claim by measuring the skewness of the distribution $N_k$ \cite{Shigeto_Hubness_2015,Article57} when projected in each direction, and the associated accuracy. We report these values in \tabref{table:hubness} for the ModelNet10 datasets. The degree of skewness is much lower when projecting the semantic feature space to the point cloud feature space, and achieves a significantly higher accuracy. This provides additional evidence that this projection direction is preferable for mitigating the problem of hubs and the consequent bias. In addition to 30 and 26 seen class settings while training with ModelNet40, we also train our proposed method using randomly selected less (20) number of seen classes, resulting in fewer data. We notice that performances decrease, but skewness scores increase while training with fewer data. It tells the overall impact of data scarcity which directly controls the generalization ability of seen features, contributes to the hubness problem and overall performance.

\begin{table}[!t]\centering
\newcolumntype{C}{>{\centering\arraybackslash}X}
\renewcommand*{\arraystretch}{1.1}
\begin{tabularx}{\columnwidth}{cccc}\hline
\# of& F2S & S2F & S2F \\ 
seen& (Inductive)  & (Inductive) & (Transductive)\\ 
 \hline
 20 & 1.43(9.91\%) & 1.08(25.77\%) & -0.06(27.42\%)\\
30 & 0.84(24.45\%) & 0.76(32.49\%) & -0.79(68.50\%)\\
\hline
\end{tabularx}
\caption{The skewness (and accuracy) on ModelNet10 with different projection directions in both inductive and transductive settings. The skewness is lower when projecting the semantic space to the input point cloud feature space, mitigating the hubness problem and leading to more accurate transductive ZSL.
}
\label{table:hubness}
\end{table}


\section{Conclusion}

With the aid of better 3D capture systems, obtaining 3D point cloud data of objects at a very large scale has become more feasible than before. However, 3D point cloud recognition systems have not scaled up to handle this large scale scenario. We apply zero-shot learning approaches to facilitate the classification of previously unseen input to readjust such a system with newly available data that have not been observed during training. We identified and addressed issues that arise in the inductive and transductive settings of zero-shot learning and its generalized variant when applied to the domain of 3D point cloud classification. We observed that in the 2D domain, the embedding quality generated by the pre-trained feature space is of a significantly higher quality than that produced by its 3D counterpart due to the vast difference in the amount of labeled training data they have been exposed to. Moreover, like ZSL on 2D images, we notice that such classification of 3D point clouds suffers from the hubness problem. The hubness problem in 3D is more severe than that observed in the 2D case. One possible reason could be that the 3D features are not trained on millions of 3D instances in the same way that 2D convolutional networks can be. In this paper, we attempt to reduce the effect of the hubness problem while performing ZSL on 3D point cloud objects by proposing an unsupervised skewness loss. In addition, we report results on Generalized ZSL in conjunction with ZSL.  Furthermore, we develop a novel triplet loss that makes use of unlabeled test data in a transductive setting. The utility of this method is demonstrated via an extensive set of experiments that showed significant benefit in the 2D domain and established state-of-the-art results in the 3D domain (both real and synthetic data) for ZSL and GZSL tasks.

\bibliographystyle{spmpsci} 
\bibliography{mybibfile}

\end{document}